# AI-Driven Predictive Modelling for Groundwater Salinization in Israel

Laxmi Pandey[a,*], Ariel Meroz[b], Ben Cheng[b], Ankita Manekar[a], Abhijit Mukherjee[c], Meirav Cohen[b], and Adway Mitra[a]

a. *Department of Artificial Intelligence, Indian Institute of Technology Kharagpur, Kharagpur, West Bengal 721302, India*
b. *Dead Sea & Arava Science Center, Mitzpe Ramon, Israel*
c. *Department of Geology and Geophysics, Indian Institute of Technology Kharagpur, Kharagpur, West Bengal 721302, India*

[*]**Corresponding author:** laxmigeophybhu@gmail.com

## Abstract

Increasing salinity and contamination of groundwater is a serious issue in many parts of the world, causing degradation of water resources. The aim of this work is to form a comprehensive understanding of groundwater salinization underlying causal factors and identify important meteorological, geological and anthropogenic drivers of salinity. We have integrated different datasets of potential covariates, to create a robust framework for machine learning based predictive models including Random Forest (RF), XGBoost, Neural network, Long Short-Term Memory (LSTM), convolution neural network (CNN) and linear regression (LR), of groundwater salinity. Additionally, Recursive Feature Elimination (RFE) followed by Global sensitivity analysis (GSA) and Explainable AI (XAI) based SHapley Additive exPlanations (SHAP) were used to estimate the importance scores and find insights into the drivers of salinization. We also did causality analysis via Double machine learning using various predictive models. From these analyses, key meteorological (Precipitation, Temperature), geological (Distance from river, Distance to saline body, TWI, Shoreline distance), and anthropogenic (Area of agriculture field, Treated Wastewater) covariates are identified to be influential drivers of groundwater salinity across Israel. XAI analysis also identified Treated Wastewater (TWW) as an essential anthropogenic driver of salinity, its significance being context-dependent but critical in vulnerable hydro-climatic environment. Our approach provides deeper insight into global salinization mechanisms at country scale, reducing AI model uncertainty and highlighting the need for tailored strategies to address salinity.

## 1. Introduction

Groundwater plays a critical role in sustaining life and development in arid environments, where surface water resources are scarce, seasonal, or unreliable. In these regions, groundwater is often the primary source of water for domestic consumption, agriculture, and industry (Tiwari et al., 2009). Its reliability, however, is increasingly challenged by both natural and anthropogenic factors that compromise not only its availability but also its quality, including salinization and contamination risks (Famiglietti, 2014; Earman & Dettinger, 2011; Howard et al., 2016). Understanding this complexity requires moving beyond mere trend identification to deciphering the causal mechanisms and interactive drivers behind water quality degradation.

Worldwide, excessive groundwater salinity is a widespread issue, being reported in more than a hundred of countries such as the Central Plains of the USA, parts of Canada, the Caribbean, Israel, Africa, India and various areas in Europe and Asia (Gurmessa et al., 2022; MacDonald et al 2016; Mondal et al., 2018; Bhakar et al., 2019; Sarkar et al., 2024). In Israel, main causes of elevation in groundwater salinity are considered to be inland encroachment of sea water, and upflow of deep-seated pressurised brine (Vengosh & Rosenthal., 1994). Traditional analytical methods including geochemical and isotopic analysis revealed Ca-chloride brines and Na-chloride saline waters as two key natural sources (Vengosh & Rosenthal., 1994). Combined analysis of high-resolution drilling with detailed hydrochemical and isotopic age dating (tritium and 14C) characterized elevated salinity. This analysis also came up with a conclusion that distribution of saline and fresh groundwater of different ages were mainly controlled by factors including separation by aquitards, connections to saline water bodies and varying flushing rates (Yechieli & Sivan., 2011). Approach of integrating physicochemical, major ion, and stable isotope analyses with geochemical and isotopic modelling revealed subsurface dissolution of soluble salts as a reason for increase in salinity (Burg et al., 2017).

The quality of groundwater is influenced by a wide range of interacting factors that can be broadly categorized as geological, climatic, and anthropogenic. The geological setting—such as aquifer structure, lithology, and proximity to saline water bodies—establishes the natural baseline of groundwater chemistry (Nosair et al., 2022; Abanyie et al., 2023). Climatic factors, including variability in precipitation, rising temperatures, and extreme events like droughts and floods, affect recharge dynamics and salinity levels (Earman & Dettinger, 2011; Amuah et al., 2023). On the human side, pressures such as over-extraction, land use change and agricultural intensification, etc have emerged as major contributors to groundwater degradation (Mosavi et al., 2021; Shtull-Trauring et al., 2022; Ghoto et al., 2025). These drivers often interact in complex, non-linear ways, producing spatially and temporally variable salinity patterns, particularly in vulnerable aquifers (Foster et al., 2021; Mukherjee et al., 2021).

Effective groundwater management depends not only on identifying threats but also on systematic, long-term observation and data collection. High-quality monitoring networks enable tracking of groundwater levels, salinity trends, recharge variability, and land use change—essential for diagnosing trends, calibrating models, and supporting decision-making (Foster et al., 2021; Amuah et al., 2023). Without robust observational data, salinization may go undetected until impacts become irreversible. To fully understand the causes and consequences of salinization, it is essential to work with high-resolution, spatio-temporal datasets. Unlike isolated measurements or short-term monitoring, long-term datasets reveal patterns and interactions that unfold over years or decades. This includes the effects of cumulative pumping, delayed responses to climate variability, and shifting land use boundaries. Such comprehensive datasets are a critical foundation for data-driven modelling efforts.

Increased groundwater salinization may be driven by both natural processes and human activities, and poses significant concerns to human and environmental health (Howard et al., 2016; Vineis et al., 2011; Mosavi et al., 2021; Shtull-Trauring et al., 2022; Grant et al., 2022). Natural causes of high salinity include rock weathering, water-rock interactions, and the evapo-concentration of salts in arid zones and predominantly in coastal zones where mixing of freshwater and seawater occurs (Abid et al., 2011; Delsman et al., 2014; Larsen et al., 2017; Villegas et al. 2018; Roy et al., 2024). Human activities, such as brine disposal from desalination plants, overpumping and salt water intrusion, mining, urban waste discharge, fertilizer application, and irrigation return flow, can intensify the issue (Mohanavelu et al., 2021; Roy et al., 2024). Increased salinity in water used for drinking and agricultural irrigation, leading to health issues like heart disease, kidney disorders, and hypertension due to the consumption of saline drinking water (Naser et al., 2017; Rosinger et al., 2025). Groundwater salinity level is usually measured in term of electric conductivity (µS/cm), total dissolved solids (TDS) (mg/L) like inorganic salts (sodium, calcium, magnesium, sulphate, and bicarbonate) and chloride levels (mg/L) (McNeill., 1988; Sahour et al., 2020; Thorsuland and Vliet, 2020; Mosaffa et al., 2021; McCleskey et al., 2023; Faghih et al., 2024; Pratiwi et al., 2024).

Traditionally physical models have been used to predict and analyse groundwater salinity; however, these models face major limitations (Khan et al., 2003; Masciopinto et al., 2017; Grunenbaum et al., 2020; Sahour et al., 2020; Schafer et al., 2020; King et al., 2022; Schiavo et al., 2024; Sarkar et al., 2024). Classical numerical and conceptual models often struggle to capture the spatial heterogeneity and non-linear interactions that characterize salinity processes at regional or national scales. These models typically rely on uncertain physical assumptions and are sensitive to boundary conditions and data gaps (Bear & Cheng, 2010). Moreover, they are time-consuming to calibrate and validate—especially in systems shaped by overlapping climatic and anthropogenic pressures (Shtull-Trauring et al., 2022; Che Nordin et al., 2021).

Since the last decade, Artificial Intelligence (AI) based Deep Learning (DL) and Machine learning (ML) has been widely used for predictions in many scientific disciplines (Schiller et al., 2025), including Geosciences (Kratzert et al., 2018; Sun et al., 2019; Tahmasebi et al. 2020; Li et al., 2021; Jiang et al 2022; Shen et al., 2023; Wang et al., 2023; Roy et al., 2025; Bramm et al., 2025; Schiller et al., 2025). Researchers are increasingly using these approaches to predict the various properties of groundwater, including salinity, and identify their drivers. ML models try to fit linear or non-linear relations between salinity and its predictors. Series of various ML models including ensemble methods like Random Forest, Gradient Boosting, conventional techniques such as Artificial Neural Network to advanced DL based convolutional neural network (CNN), Long-short term memory (LSTM) models have been found to show significant success in understanding the complex datasets and make predictions based on a set of predictor variables (Banerjee et al., 2011, Alagha et al., 2017, Lal and Dutta, 2018; Sahour et al., 2023; Mosavi et al., 2020 and 2021; Boudibi et al., 2024; Roy et al., 2025). However, these models may not be necessarily suitable for identifying which predictors have causal influence on the target variable (in this case, salinity) (Shmueli, 2010; Mullainathan and Spiess, 2017). Thus, to overcome this limitation of ML method, in this study we have used double or debiased machine learning for causal hybrid modelling. The fundamental promise of Double Machine Learning (DML) lies in its potential to leverage flexible machine learning techniques to effectively account for observed confounding variables (Jung et al., 2021). While performing causal inference for groundwater salinization using DML, the focus is primarily on estimating the causal effects of various features on salinity levels while simultaneously controlling for confounding variables. This approach improves our understanding of how specific interventions or environmental changes influence groundwater quality, thereby supporting the development of effective management strategies (Chernozhukov et al., 2016; Cohrs et al., 2024; Fuhr et al., 2024).

In recent years, hydrological modeling has undergone a renaissance through Explainable AI (XAI), which reveals decisive features, quantifies their contributions, and tests model robustness (Jiang et al., 2022; Ghaffarian et al., 2023; Pradhan et al., 2023; Topp et al., 2023; Wu et al., 2023; Jung et al., 2024; Ye et al., 2025; Bramm et al., 2025). These methods work by integrating the AI models, which have learned patterns from hydrological data, with analytical techniques that extract explanatory insights. This allows hydrologists to understand the learned relationships, for example, by identifying the relative importance of different factors affecting water quality (Maier et al., 2024). The common XAI methods include: Anchors Explanations (Ribeiro et al., 2018), LIME+Anchors (Molnar, 2020), Shapley Values (Lundberg and Lee, 2017), Integrated Gradients (Sundararajan et al., 2017), LIME (Ribeiro et al., 2016), Wachter et al. (Wachter et al., 2017), DiCE or Diverse Counterfactual Explanations (Mothilal et al., 2020), CEM or Contrastive Explanations Method (Jacovi et al., 2021) and Counterfactual Latent Representations (Dhurandhar et al., 2018). However, critical aspect of these methods is that they shouldn't be applied blindly as XAI is useful only if applied to well-developed, reliable AI models (Maier et al., 2024). Recent advances in XAI now enable a shift from black-box modelling to transparent, interpretable analysis thereby avoiding biased models (Samek et al., 2021; Bramm et al., 2025). Tools such as SHAP (Shapley Additive Explanations) and global sensitivity methods (Sobol, Morris, 2001) allow researchers to identify, quantify, and rank the predictor variables, based on their influence on the target variable (Linardatos et al., 2020; Molnar, 2020; Stein et al., 2022; Saltelli et al., 2004; Francom & Nachtsheim, 2025; Panigrahi et al., 2025). This interpretability is not only scientifically valuable—it is increasingly recognized as essential for real-world decision-making. As emphasized in UNESCO's recent report on Artificial Intelligence for Water Management (2023), transparency and trust in AI systems are critical for their adoption in environmental governance and water policy. The report highlights that explainable AI can bridge the gap between technical innovation and practical application, particularly in risk-sensitive contexts like groundwater salinization, where decisions must be justified to regulators, stakeholders, and the public. By aligning with these global recommendations, the current study contributes not only to the advancement of groundwater modelling, but also to the broader agenda of responsible and actionable AI in water management. Despite the significance of this environmental issue of elevated groundwater salinity in water resources, yet there is a lack of comprehensive studies examining the variability and factors influencing groundwater salinity at regional level.

In order to address this gap, this study intends to predict the occurrences of rising groundwater salinity across Israel, defined in terms chloride ($Cl^-$) concentrations, leveraging field observations and using ML models and XAI methods. The research is carried out at three stages : (1) firstly, causality in the dataset is recognised from double machine learning (DML), (2) secondly, we predict groundwater salinity using various ML based regression models—random forest (RF), eXtreme Gradient Boosting XGBoost, linear regression (LR), Long Short-Term Memory (LSTM), Feed Forward Neural Network (FFNN) and convolution neural network (CNN), and (3) thirdly, using XAI we go over various ML based regression models to show their strength and ability to explain the predicted salinity. Specifically, XAI is employed here to interpret salinity dynamics at three levels: (a) the relative importance of input variables; (b) the direction and strength of each variable's influence; and (c) the interactions between variables.

The novelty of this work lies in establishing an XAI-centric analytical framework to not only predict groundwater salinity but, also more importantly, to explain its driving mechanisms, thereby moving from correlation to causality and from prediction to understanding. The XAI approach used in the study provides both scientifically grounded insights and practical tools for managing groundwater quality in complex, data-rich environments. While many studies have modeled salinity risk using ML (e.g., Mosavi et al., 2021; Nosair et al., 2022; Barzegar & Asghari, 2016), few have constructed such a multi-layered interpretive framework. By making XAI explanations central to our methodology, we bridge the gap between data science and domain science, creating models that are both powerful and capable of informing the sustainable management of precious groundwater resources in Israel and analogous regions worldwide facing salinization threats. The contribution of this work is therefore defined by its rigorous application of XAI to decompose model predictions, validate them against hydrological domain knowledge, and generate actionable, transparent insights for water resource management.

## 2. Material and Methods

*2.1. Study Area*

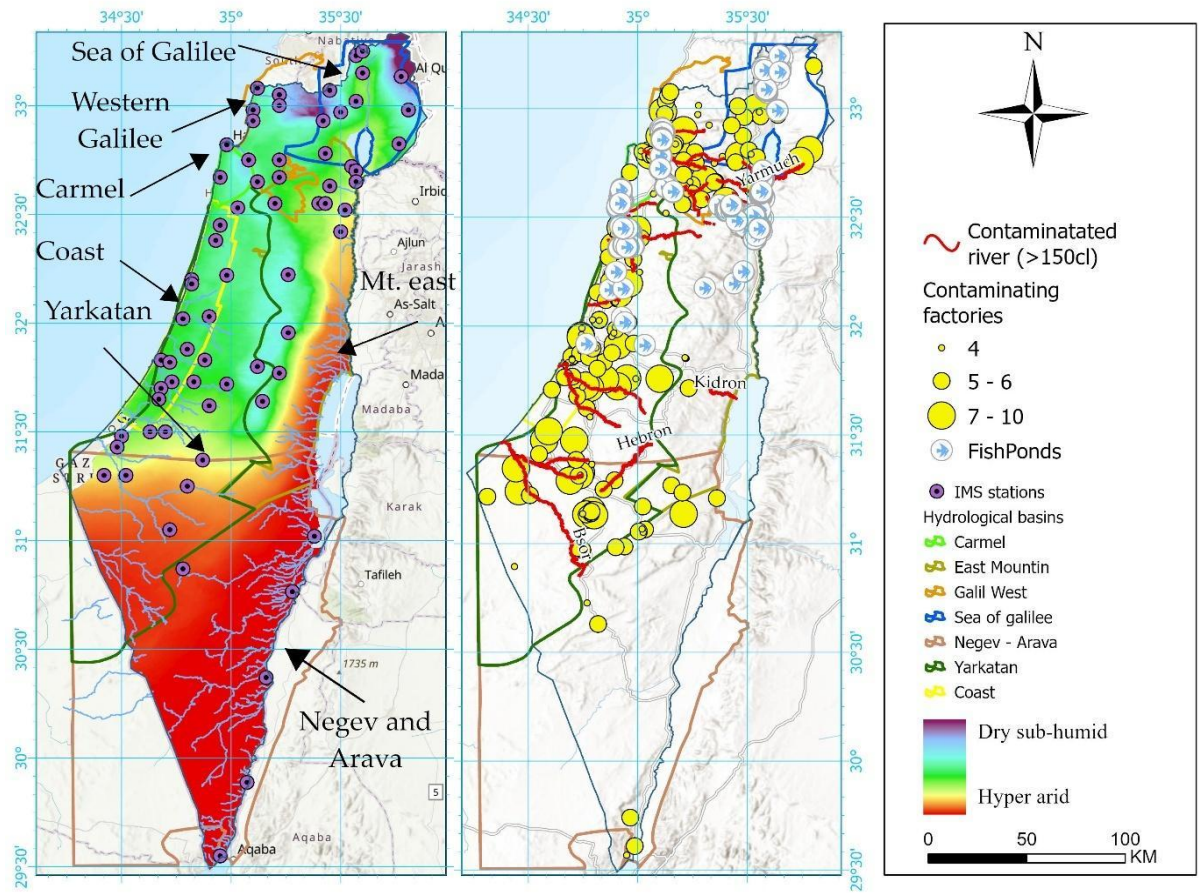

Fig. 1. Map of hydrogeological basins across Israel. Note that some basins partially overlap due to shared geological and hydrological boundaries.

This study focuses on groundwater salinity, specifically chloride concentrations, across Israel's major hydrological basins, namely the Coastal basin, Yarqon-Taninim, Western Galilee, Mount Carmel, Sea of Galilee, Eastern Mountain, and Negev and Arava (see Fig. 1). The analysis concentrates on the shallow (upper) aquifers within these basins, as these systems are in direct contact with atmospheric conditions and are thus more responsive to changes in climate, land use, and surface inputs.

The country is located on the eastern bank of the Mediterranean Sea, and is characterized by a broad range of climatic, geological, and anthropogenic conditions. Its national groundwater system spans seven primary hydrological basins, each exhibiting distinct lithological structures, recharge regimes, water management practices, and varying degrees of vulnerability to salinization. Climatically, Israel lies along a pronounced aridity gradient, spanning from hyper-arid zones in the south, through arid and semi-arid regions, to small areas of dry sub-humid climate in the north. The country is also marked by a dense population (386 inhabitants per km²), high living standards, and a highly developed agricultural and industrial sector. Though water is also the basis for some of the cross boundaries cooperations in this area like the water agreement with Jordan. Its complex geopolitical environment often limits regional cooperation in managing shared environmental and water challenges. Furthermore, Israel encompasses diverse geological and morphological zones—from coastal plains prone to seawater intrusion to inland basins with low natural recharge and intensive anthropogenic stress (Yechieli & Sivan., 2011).

## 2.2. Dataset Preparation

Groundwater salinity is influenced by a wide range of factors, which may be grouped into three broad categories: physical (geological and geographical), climatic, and anthropogenic. The anthropogenic drivers are further divided into: (1) operational water system factors, including groundwater extraction and artificial recharge, and (2) contamination sources and hotspots such as pollution originating from different land uses. The dataset compiled for this study integrates long-term observational records from various Israeli governmental authorities alongside global datasets, resulting in consistent, high-resolution spatial and temporal coverage. Wherever possible, we constructed a yearly time series for each variable. However, physical parameters—such as lithology, elevation, or distance from major rivers—were treated as temporally static, given their stability over time. For some anthropogenic variables where regular data collection was unavailable (e.g., point sources of pollution), we relied on spatial proxies such as the presence or location of infrastructure, or used multi-year averages of salinity-related values reported in recent monitoring records. Groundwater salinity observations (Cl (mg/L)) were obtained from the Israeli Hydrological Service, which maintains a national groundwater monitoring network. Each borehole includes a time series of salinity measurements, as well as its spatial coordinates, and contextual hydrogeological information.

To prepare a comprehensive and temporally aligned dataset where each borehole-year pair was enriched with the relevant set of predictor variables across all three domains (physical, climatic, anthropogenic), we considered each borehole location as a spatial anchor point, linked to the environmental and anthropogenic variables via: (a) raster data (e.g., temperature, land cover, elevation), extracted based on the pixel value at the borehole coordinates, and (b) vector data (e.g., location of fishponds, industrial facilities, river proximity) linked spatially with the boreholes using buffer zones defined for each variable in the Israeli water system, applied uniformly across all boreholes. The dataset was organized into a tabular format with each row representing a borehole–year observation comprising the target variable (chloride concentration) and associated predictor variables. A complete list of all variables used in the analysis is presented in Table 1. These variables were selected based on literature review, hydrological relevance, data availability, as per the research objectives.

**Table 1: List of selected predictor variables of groundwater salinity with their respective sources for Israel**

| Category | | Variables | Time resolution | Data Source |
|---|---|---|---|---|
| Geographical | Topography | Topographic Wetness Index (TWI) [4m/pixel] | Constant | MOAG (Ministry of Agriculture) |
| Geographical | Hydro-Geology | Distance to rivers [km] | Constant | Survey of Israel |
| Geographical | Hydro-Geology | Distance to shoreline [km] | Constant | Survey of Israel |
| Geographical | Hydro-Geology | Distance to saline bodies [km] | Constant | NPA (Nature protection authority) |
| Geographical | Soil | Soil texture (clay, silt, sand) [%] | Constant | MOAG |
| Climatic (Natural) | Temperature [$^0$C] | Maximum Temperature | Yearly | IMS (Isreal Meteorological Service) |
| Climatic (Natural) | Temperature [$^0$C] | Minimum Temperature | Yearly | IMS |
| Climatic (Natural) | Precipitation [mm] | Yearly sum | Yearly | IMS |
| Climatic (Natural) | | Aridity Index | Yearly | |
| Anthropogenic | Water System Operation | Water extraction [MCM/year] | Yearly | IHS (Israeli Hydrological Service) |
| Anthropogenic | Water System Operation | Water Induced (artificial recharge) [MCM/year] | Yearly | IHS |
| Anthropogenic | Water System Operation | Absolute pumping depth (drill topographic elevation | Constant | IHS |

| | | | - pumping point elevation) [m] | | |
|---|---|---|---|---|---|
| | | | Land uses (classified into Build-up/ disrupted, agricultural, water, natural and planted forest) | Yearly | https://maps.elie.ucl.ac.be/CCI/viewer/ |
| | | | Population density [people/km$^2$] | Yearly (1998-2020) | https://landscan.ornl.gov |
| | | Pollution Sources | Fishponds [No./km$^2$] | Constant (Based on 2020) | MOAG |
| | | | Polluted rivers Cl>150 [mg/l] in 1km$^2$ | Constant (Based on 2020) | MoEP (Ministry of Environmental Protection) |
| | | | Industrial factories -3,7-10 Cl LOG order [No./km$^2$] | Constant (Based on the average value 2014-2020) | MoEP |
| | | | Industrial farms [No./km$^2$] | Constant (Based on 2020) | MOAG |
| | | | Area of TWW irrigated agricultural fields in 1km$^2$ [km$^2$] | Transformed into an annual basis, based on surveys conducted in the years 2006, 2010, 2012, 2014, 2016, 2018, 2020 | NPA |
| | | | TWW Cl levels [mg/l] in 1km$^2$ | Transformed into an annual basis, based on surveys conducted in the 2006, 2010, 2012, | NPA |

| | | | 2014, 2016, 2018, 2020 | |

*2.3. Model Workflow*

   a. Data Preprocessing:

We prepared the raw data to be fed into the model after bringing all input features at comparable scale. In the literatures it has been reported that values of cl above 5000 mg/l are considered saline (below its moderately saline) and in Israel above 4000 mg/l is considered saline (Hong et al., 2023). Thus, the target salinity dataset was at first filtered to remove the values bigger than 4000 mg/L as outliers. The target salinity variable was then used in its original continuous form for the further analysis. Variance thresholding and correlation analysis is used to remove highly correlated features (threshold > 0.95), while missing values are handled by mean imputation for numerical columns. Text categories are converted into numerical format via one-hot encoding. Since, different hydrological basins have different numbers of drills (samples). This causes geographical imbalance and as a result of which the model is biased toward the basins with higher density. So, to overcome this bias in the model, we have employed an inverse square-root density weighting approach. Sampling has been done per drill and basins with lower drill densities receive higher weights to balance their influence during model training and vice-versa, preventing regional bias. Each basin is assigned a normalized weight based on its provided density value, with low-density basins receiving higher weights than high-density basins to ensure fair representation of drills from all basins. These basin-level weights are applied to all individual drills within each basin, ensuring that samples from underrepresented regions have proportionally greater impact on model training.

b. Causality Analysis using Double Machine Learning:

After preprocessing of the dataset firstly, we carried out causality analysis using a technique called Double Machine Learning method which is capable of delivering unbiased estimates in situations where there are many variables 'X' out of which only a few have a causal impact on a target variable 'Y'. The other variables may still have high correlation with the target variable, but their relation need not be causal, and they are considered as 'confounders'. In each step of Double Machine Learning, the main aim is to conclude if a particular variable T (referred to as 'treatment') has a causal impact on variable Y (referred to as target or outcome), in presence of other variables 'X' which may have association with either of T and Y. The formulation of the Double ML method consists of three main steps: (a) ML models such as Ordinary Least Square (OLS), Random Forest, Neural Networks, XGBoost are used to predict the outcome variable (Y) as a function of the covariates (X). The goal is to capture the relationship between (Y) and (X) irrespective of treatment T. (b) Similarly, we use ML models to predict the treatment variable (T) as a function of the covariates (X). This step aims to capture the relation between the treatment T and the covariates X. After estimating the models for both the outcome and treatment, the residuals from these models are computed. These residuals represent the information contained in them which is not attributable to covariates and the variation in outcome not explained by covariates alone. (c) Finally, we perform an OLS regression of the residual outcome on the residual treatment. This regression estimates the causal effect of the treatment on the outcome, independent of the confounding effects of the covariates. The coefficient obtained from this regression represents the estimated causal effect of the treatment on the outcome.

   c. Training/Evaluation and Performance:

Next, the model is made to learn using the training dataset and its performance is objectively measured on validation dataset (validation set) using standard performance metrics like MAE, RMSE, and $R^2$ to quantify prediction error and accuracy. We used various machine learning based regressions models defined below including LR, RF, XGBoost, ANN, LSTM and CNN. The general explanation of these models is given in Appendix A.

The feature importance is evaluated and the best-performing model is selected based on cross-validated $R^2$ score, followed by XAI analysis to interpret feature importance.

   d. Feature Importance Analysis:

Various methods like recursive feature elimination (RFE) infused with Random Forest were used to evaluate the feature importance followed by Explainable AI (XAI) techniques such SHAP (SHapley Additive exPlanations) and Sobol-based Global Sensitivity Analysis (GSA) applied to interpret the model's predictions, identifying and ranking the input features which were found to be most influential in the decisions making process. These methods are explained in Appendix B.

The workflow of the methodology used for groundwater salinity prediction across Israel is shown below in Fig. 2

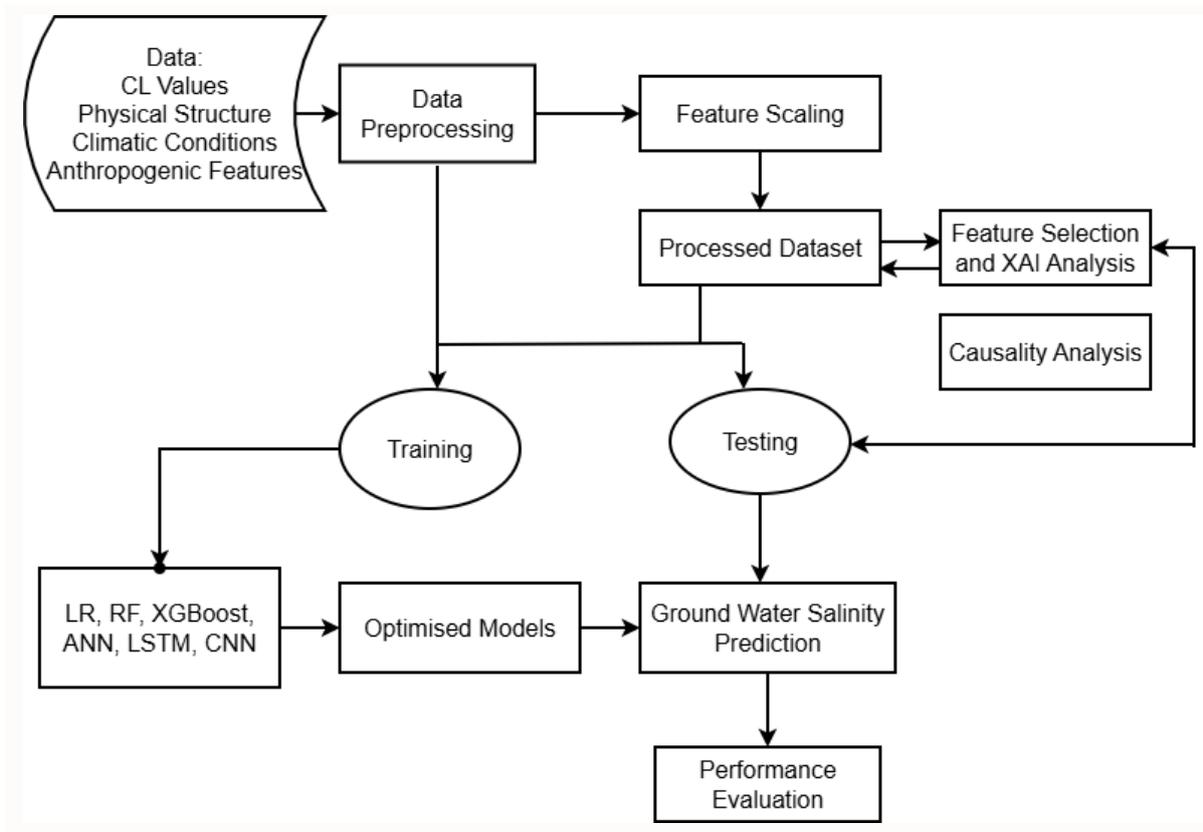

Fig. 2. Workflow employed for groundwater salinity prediction

## 3. Results and Discussion

In the study area above, we used Inverse Square-Root density weighing approach to deal with geographical imbalance across hydrological basins and avoid model's biasness. This weighting strategy is implemented consistently across all machine learning models (CNN, ANN, LSTM, XGBoost, Random Forest, and Linear Regression) through their sample-weight parameters, complementing prior feature-level imbalance handling that included removing constant features, eliminating highly correlated features, and standard scaling. The approach ensures geographical fairness without altering the original dataset, allowing all basins to contribute meaningfully to the model regardless of their sample size while maintaining the integrity of temporal validation splits across drill sites.

*3.1. Double Machine Learning (Causality analysis)*

Double ML assigns a score to different predictor variables. These scores can be interpreted as the causal relation between the predictor and the predictand, i.e. groundwater salinization. Based on the DML analysis, Fig. 3. reveals a clear hierarchy of statistically significant predictors (p value < 0.05 as cutoff) having varying magnitudes of causal influence on groundwater salinity, with both natural hydrogeological conditions and human activities playing crucial roles. Features shortlisted from different predictive models having causal relation with groundwater salinity, include distance to saline body, distance from river, drill depth, precipitation, shoreline distance, TWI, TWW Cl level, TWW irrigated area, temperature, distance to agriculture field.

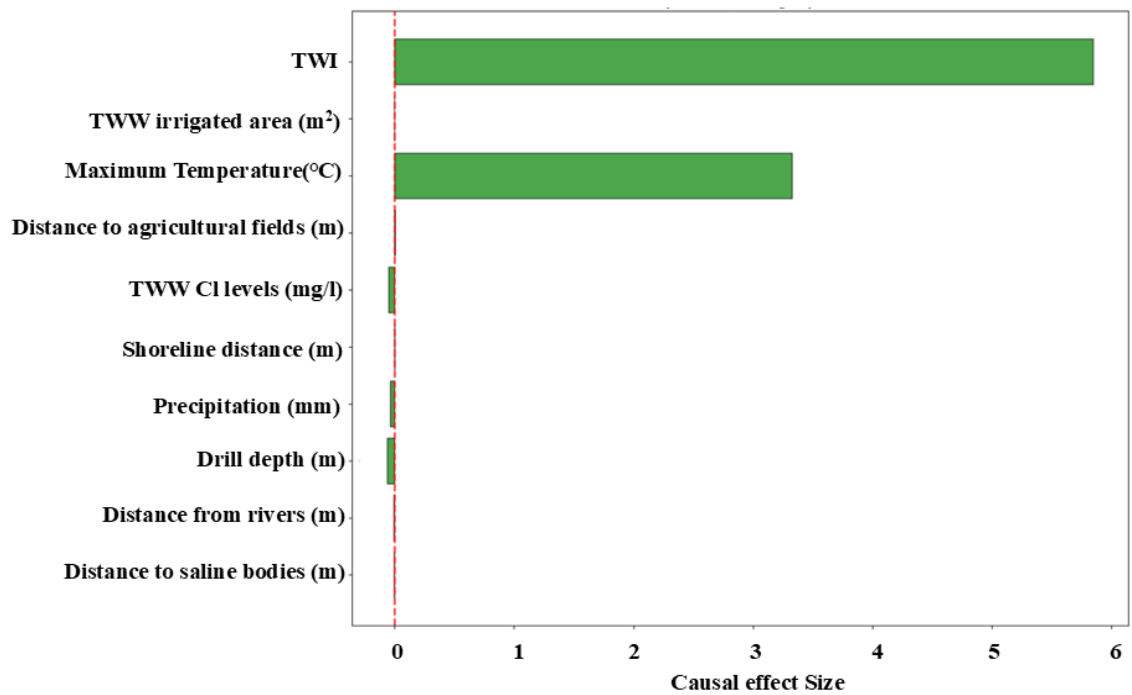

Fig. 3. Predictors having causal effect on groundwater salinity.

Topographic influences in terms of Topographic Wetness Index (TWI) can be seen as the strongest predictor (causal effect: ~5.8), indicative of landscape position and water accumulation patterns having primary hold on salinity distribution. This substantial positive effect depicts that region with higher wetness indices (mainly convergence zones and valley bottoms) experience significantly elevated salinity levels, possibly due to the accumulation of dissolved salts through surface and subsurface water flow pathways (Corwin et al., 2007; Meles et al., 2020).

Among the climatic drivers, Maximum temperature emerges as the second-strongest positive driver (causal effect: ~3.3), highlighting the role of climatic aridity and evaporation in concentrating salts in groundwater. Conversely, precipitation showing a notable negative effect (causal effect: ~-0.036) further supports the importance of freshwater inputs, where increased rainfall actively reduces salinity through dilution and aquifer flushing.

Notably, several drivers show negligible causal effects in this analysis. Distance to saline bodies (~-0.004), distance from rivers (~-0.006), shoreline distance (~0.002), and TWW irrigated area (~5.72E-06) all have effect sizes near zero. This indicates that, in this system, proximity to surface saline features, rivers, or the coast, as well as the spatial extent of treated wastewater irrigation, do not substantially directly influence groundwater salinity patterns.

Interestingly, though very slight, positive effect is observed for distance to agricultural fields (~0.005). This minimal relationship suggests a weak tendency for salinity to be slightly lower closer to fields, potentially suggesting localized dilution from irrigation return flows, though the effect is not strong. As excess freshwater irrigation percolates through agricultural soils, it recharges underlying aquifers with lower-salinity water, effectively reducing groundwater salinity in the immediate vicinity of agricultural areas. This return flow mechanism overprints natural salinity patterns and demonstrates how human agricultural practices can actively improve local groundwater quality through managed freshwater inputs (Rosenthal et al., 1992; Yechieli et al., 1992; Oren et al., 2004; Chefetz et al., 2008; Bekkam et al., 2013; Avisar and Eliraz, 2019; Zhang and Shen, 2019; Hashem et al., 2021; Klapp et al., 2023).

DML analysis emphasizes TWI and maximum temperature emerged as the dominant variables, indicating that—after accounting for the influence of the remaining covariates the residual, independent signal in annual groundwater salinity is most consistently associated with (i) local topographic controls on moisture accumulation and flow–recharge realization (TWI), and (ii) the energy balance and concentration potential linked to temperature (e.g., evaporative concentration and temperature-dependent hydrogeochemical processes).

Next, after causality analysis we report the performance of the different Machine Learning models employed for prediction of groundwater salinity.

*3.2. Groundwater salinity predictions*

The performance metrics for regression by the different models (linear regression, random forest, XGBoost, ANN, LSTM and CNN) are shown through scatter plots in Fig. 4.

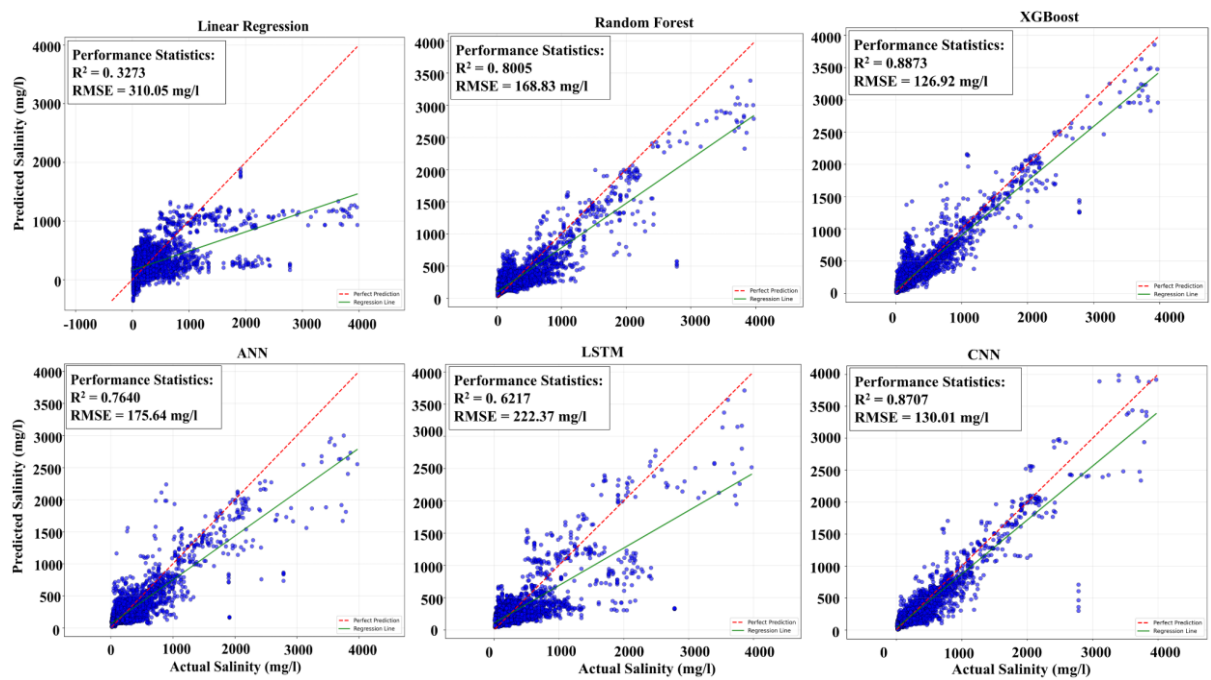

Fig. 4. Performance of different regression-based ML models in terms of scatter plots from models: (a) Linear regression, (b) Random Forest, (c) XGBoost, (d) ANN (e) LSTM and (f) CNN

The analysis of performance of six machine learning models for predicting groundwater salinity reveals distinct patterns in their predictive capabilities. It is found that XGBoost emerged as the most effective model with an exceptional $R^2$ of 0.887, demonstrating accuracy with the lowest root mean square error (126.92 mg/l) and mean absolute error (69.36 mg/l), followed by CNN ($R^2$ = 0.871) which effectively captured spatial-temporal patterns in the data. Random Forest delivered solid performance ($R^2$ = 0.801) with good generalization, while ANN showed moderate results ($R^2$ = 0.764), suggesting potential for architectural optimization. Interestingly, LSTM underperformed ($R^2$ = 0.622) despite its temporal modelling capabilities, indicating limited sequential dependencies in the current dataset, whereas Linear Regression proved fundamentally inadequate ($R^2$ = 0.327), confirming the complex, nonlinear nature of salinity processes that cannot be captured by simple linear relationships.

The clear supremacy of tree-based ensembles like XGBoost over both traditional statistical methods and complex neural architectures highlights the importance of feature interactions and hierarchical decision boundaries in modelling groundwater salinity dynamics, providing valuable guidance for selecting appropriate modelling approaches in hydrological applications.

Further, we discuss the importance of features for the different predictive models using the aforementioned methods.

### 3.3. Feature Importance Analysis

We evaluated feature importance across multiple models including LR, RF, XGBoost, LSTM, ANN, CNN. While the ranking of features may vary between models, our objective is to identify a core set of predictors with stable importance rankings.

Recursive feature elimination (RFE) was used to evaluate the feature importance. The plot in Fig. 5 illustrates the hierarchy of the set of features identified using RFE method, where natural geogenic factors dominate, but anthropogenic activities also play a measurable role in salinity distribution. The influential drivers identified from RFE-RF method included precipitation, distance to saline body, temperature, drill depth, distance from rivers, aridity index, TWW Cl levels and TWW irrigated area.

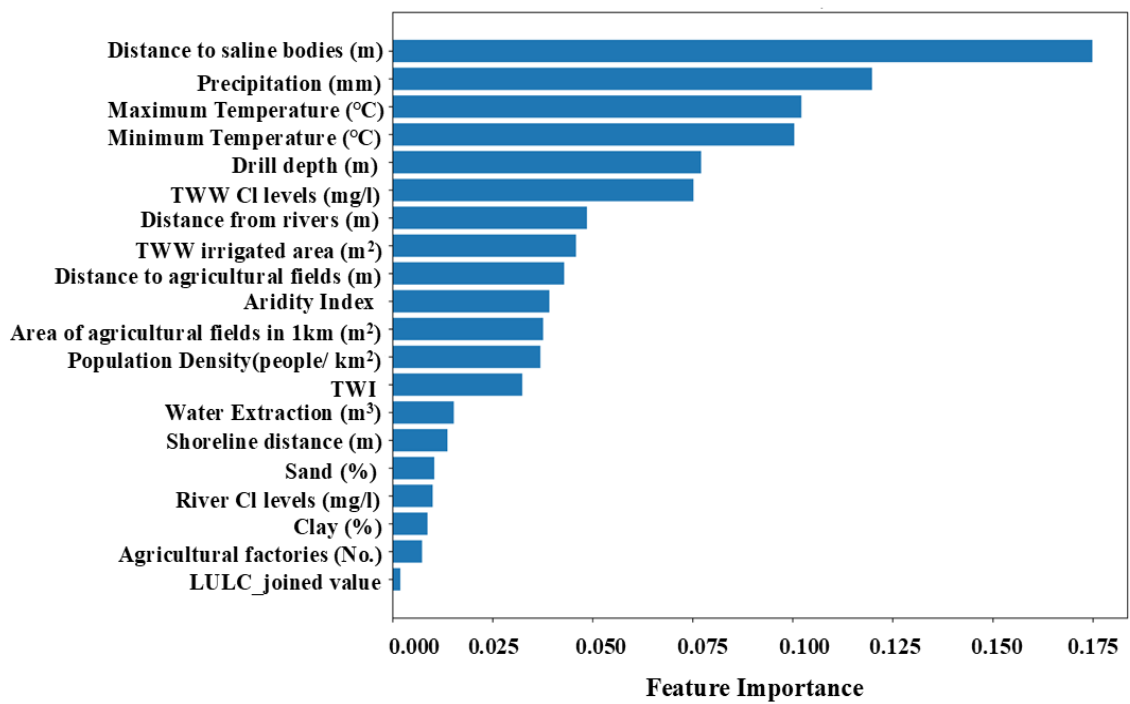

Fig. 5. RFE based feature importance of influential predictors affecting groundwater salinity.

Distance to saline bodies is the top most important predictor of groundwater salinity, indicating that proximity to natural saline sources is the dominant controlling factor. Following this, climatic drivers like precipitation and temperature are next highly important, highlighting the role of dilution and evaporation-concentration processes. The occurrence of human-influenced predictors including TWW Cl levels and TWW irrigated area in the list of top features assures that reuse of treated wastewater significantly influences salinity. Whereas, hydrogeological controlling drivers such as Drill depth and distance from rivers reveal the significance of aquifer characteristics and freshwater proximity. Many of these rivers are ephemeral rivers transporting saline water thereby increasing the salinity which can also be seen in the SHAP analysis

**GSA or** Sobol based Global Sensitivity Analysis (GSA) was used to assess the variance in models' salinity predictions. Among the different model runs (shown in Fig. 6 and appendix), GSA analysis from XGBoost, CNN and RF model show the most consistent patterns in feature sensitivity ranking thereby revealing a stable core of dominant drivers while highlighting important methodological insights.

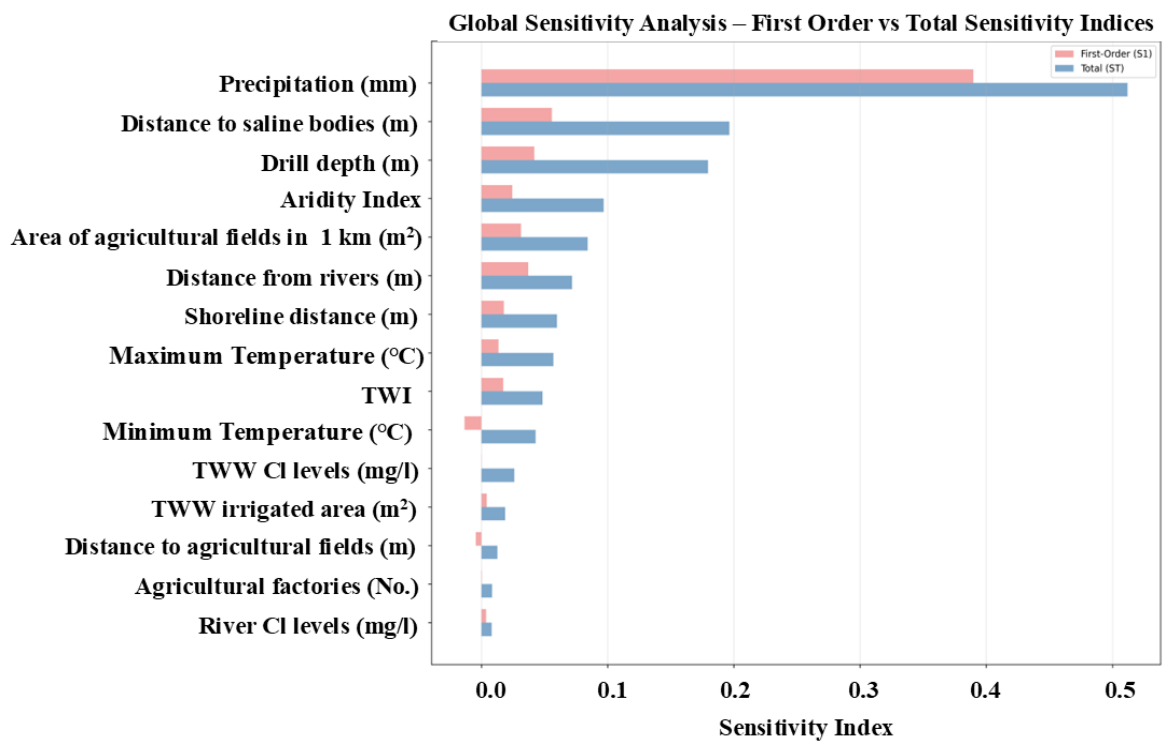

Fig. 6. XGBoost-based Sobol Indices from Global sensitivity analysis (GSA).

Precipitation and Distance to saline bodies can be seen as the two most sensitive parameters, with precipitation typically showing the highest first-order sensitivity ($S_1$) index around 0.38, indicating its strong individual effect on salinity predictions. Following these primary drivers, a Drill depth (m), Aridity index, and Area of agricultural fields maintain relatively stable moderate sensitivity across analyses. The consistency extends to the clear separation between first-order and total sensitivity indices, mainly for precipitation and saline body proximity, suggesting that these features drive salinity through both individual effects and interactions with other parameters. However, even in these most stable results, features like temperature parameters and agricultural indicators show noticeable ranking variations, underscoring that while GSA reliably identifies the dominant hydrological and climatic drivers, it remains less stable for quantifying the relative importance of secondary anthropogenic and soil-related factors in groundwater salinity modelling.

**SHAP -** Further to report the limitations of the above-mentioned methods, we computed Shapley values to quantify the contribution of each feature to a certain prediction. The most significant results of SHAP analysis, shown in Fig. 7, yielded significant insights into the covariates including precipitation, distance to saline bodies, drill depth, distance from rivers, area of agriculture field, temperature, shoreline distance, aridity index which affected the target variable groundwater salinity. The consistency of SHAP based across different ML models is showcased through bar chart in Fig. 8, while the individual SHAP summary results obtained from each of the models is shown in Appendix C:

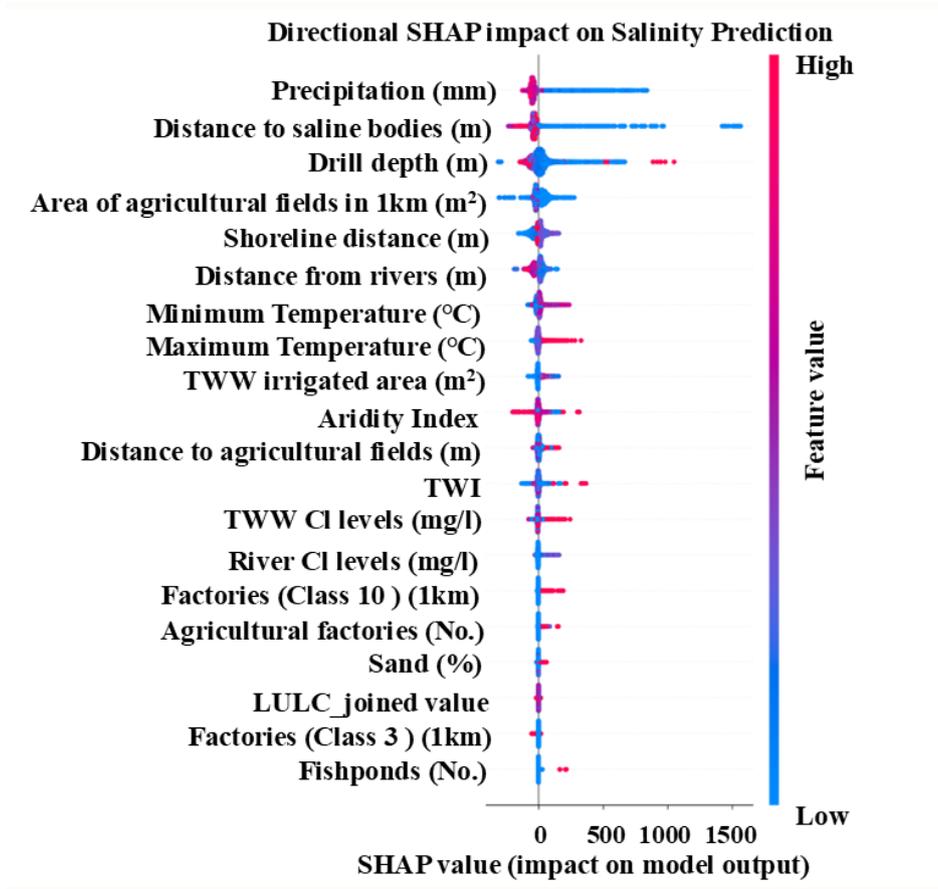

Fig. 7. XGBoost-based SHAP summary plot.

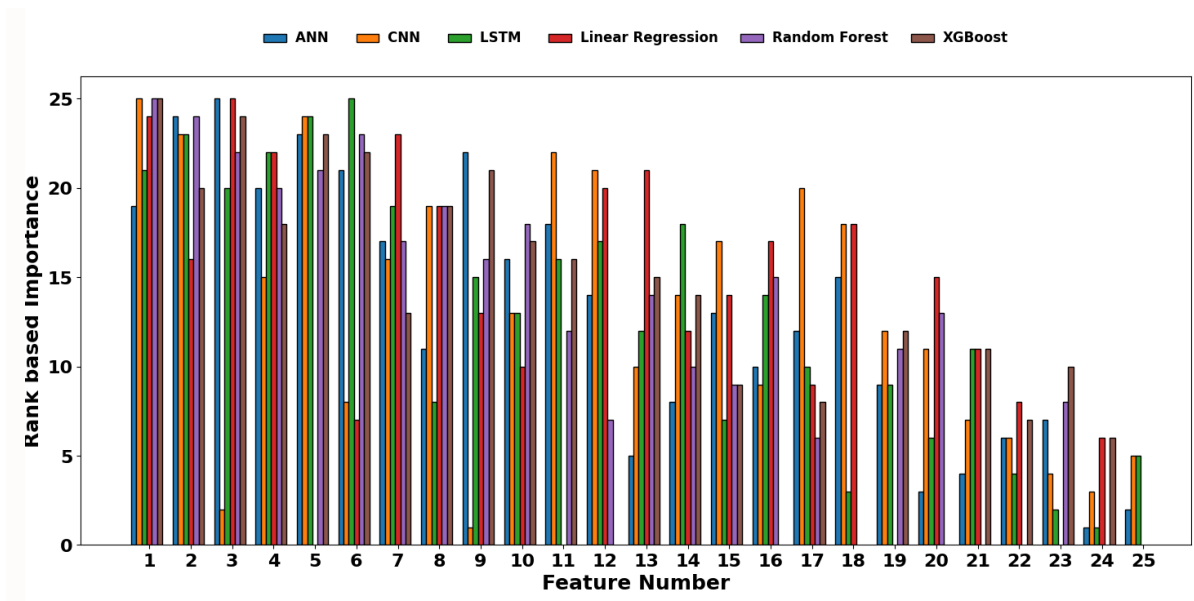

Fig. 8. Comparison of SHAP based rank from ML based ANN, CNN, LSTM, LR, RF and XGBoost regression model. The figures depicts that most influential features have higher bars. The numbers on x axis define the features or predictors including (1) Precipitation (mm), (2) Distance from rivers (m), (3) Distance to saline bodies (m), (4) Maximum Temperature (°C), (5) Drill depth (m), (6) Area of agricultural fields ($m^2$), (7) TWW CL levels (mg/l), (8) Minimum Temperature (°C), (9) Shoreline distance (m), (10) TWW irrigated area ($m^2$), (11) Aridity Index, (12) Clay (%), (13) Distance to agricultural fields (m), (14) TWI, (15) Sand (%), (16) Population density (people/$km^2$), (17) LULC, (18) Silt (%), (19) River Cl level (mg/l), (20) Water extraction ($m^3$), (21) Factories (Class 10) (No.), (22) Factories (Class 3) (No.), (23) Agricultural Factories (Class 10) (No.), (24) Fishponds (No.) and (25) Factories (Class 7) (No.).

The SHAP analysis reveals a complex summary of climatic, hydrogeological, and anthropogenic factors driving groundwater salinity patterns. Climatic driver like precipitation is seen to emerge as the most influential predictor validating the fact that low rainfall values have an excessively large positive impact on the predicted salinity, indicating that drought conditions (indicated by low SHAP values of precipitation) are a primary driver of high salinity events. Conversely, beyond a certain threshold, increased rainfall yields diminishing returns in salinity reduction. This demonstrates that while freshwater recharge is critical for dilution, the absence of it is an even more powerful factor concentrating salts, highlighting a key vulnerability to climate fluctuations. Temperature parameters (both maximum and minimum) show notable consistent and positive relationship with predicted salinity, where higher temperatures correspond to higher SHAP values, pushing predictions toward increased salinity. It indicates that higher temperatures likely exacerbate salinity through increased evaporative losses, reducing effective moisture and concentrating salts in the aquifer.

Next, hydrogeological factors including proximity to saline bodies and drilling depth demonstrate significant impacts, revealing that natural saline sources and deeper aquifer layers contribute substantially to groundwater salinity. The substantial influence of agricultural field area within 1km radius highlights the importance of land use patterns, while shoreline distance and distance from rivers indicates that shorter distances to the shoreline correlates with higher salinity and this correlation is attributed to the influence of the eastern saline aquitard. Near, the coast, the salty aquitard allows the salt to slowly seep into the freshwater aquifer thereby increasing the salinity. Whereas proximity to ephemeral rivers increases salinity, as they act as a conduit or a pathway, allowing the contaminated water to infiltrate and recharge the groundwater system rapidly, before it can be filtered. As a result, groundwater near these ephemeral rivers shows higher salinity due to this direct, contaminated recharge.

Anthropogenic drivers including TWW related variables (irrigated area and chloride levels) appear in the middle-lower tier of importance, suggesting that while wastewater reuse contributes to salinity, its impact is secondary to dominant natural hydrogeological factors. Although where several localized studies have examined the issue of TWW influence on groundwater salinity, some have reported no adverse effects of TWW irrigation on groundwater. While others identified groundwater contamination linked to such practices (Zhang et al., 2019; Hashem et al., 2021; Klapp et al., 2023), yet the overall impact of TWW on groundwater systems remains insufficiently understood. Comprehensive, large spatio-temporal scale, assessments are still lacking, yet understanding contaminant pathways and persistence in subsurface environments is essential for evaluating the long-term sustainability and safety of TWW reuse, particularly in groundwater-dependent regions.

The analysis (Fig. 7 and 8) also reveals few prominent drivers whose SHAP trend is not clearly visible in XGBoost models but can be clearly seen in other model derived SHAP results like CNN. These drivers include TWW irrigated area ($m^2$) and Distance to agricultural fields (m). This hierarchy of influences emphasizes that groundwater salinity is primarily governed by natural hydrological processes—particularly precipitation patterns and riverine interactions—with human activities including agriculture and wastewater management introducing secondary but meaningful modifications to the natural salinity regime. The results highlight the need for integrated water resources management that addresses both the dominant climatic drivers and the measurable anthropogenic contributions to groundwater quality degradation.

*3.4. Feature Importance Hierarchy: SHAP, GSA, and RFE Agreement Analysis*

This hierarchical visualization in Figure 9 presents a comparative analysis of feature importance rankings derived from three distinct methods including SHAP (SHapley Additive exPlanations), GSA (Global Sensitivity Analysis), and RFE (Recursive Feature Elimination) applied to an XGBoost model for groundwater salinity prediction. While spearman correlation coefficient between RFE-SHAP-GSA based ranks from different models runs (LR, RF, XGBoost, ANN, LSTM and CNN) is given below in Table 2.

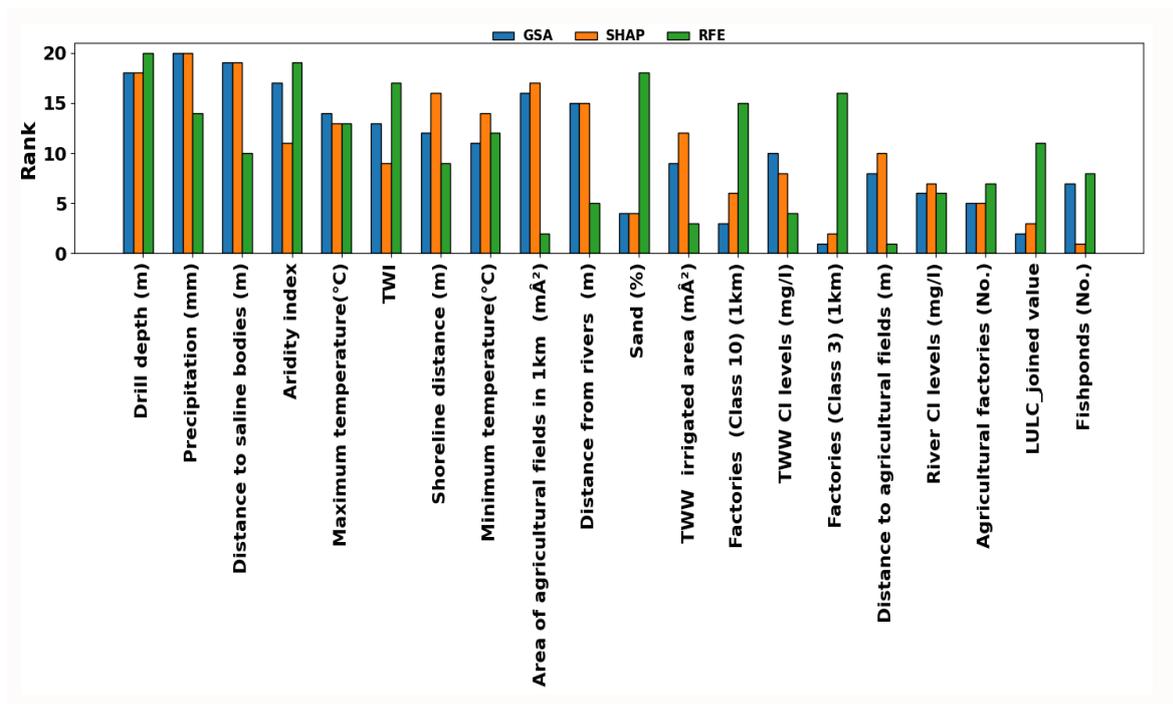

Fig. 9: Comparison of RFE, SHAP and GSA based feature ranks from XGBoost regression model.

Table 2: Spearman rank correlation between RFE-SHAP-GSA for different ML models.

| Models | RFE-SHAP | SHAP-GSA | GSA-RFE |
|---|---|---|---|
| Linear Regression | 0.76 | 0.76 | 1 |
| Random Forest | 0.85 | 0.80 | 0.76 |
| eXtreme Gradient Boost (XGBoost) | -0.04 | 0.89 | 0.08 |
| ANN | -0.16 | 0.05 | 0.73 |
| LSTM | 0.45 | 0.38 | 0.49 |
| CNN | 0.64 | -0.12 | -0.14 |

For hydrogeological and climatic drivers including drill depth and distances to saline bodies, distance from rivers, and shoreline distance, strong agreement can be seen among SHAP, GSA, and RFE methods. It indicates that these drivers are inherent structural controls whose importance is robust and consistently detectable regardless of analytical approach. Climatic drivers, such as precipitation and temperature variables, show strong inter-method agreement. In contrast, anthropogenic drivers related to TWW, including chloride levels and irrigated area, appearing in middle-lower tiers depict secondary importance to above drivers. It signifies that TWW impacts are not universal drivers but conditional influencing drivers as their importance emerges only in specific hydro-climatic contexts, through non-linear interactions, or beyond certain thresholds. Thus, the figure underscores a critical insight for management: while hydrogeological and climatic factors require systematic and region-wide attention, anthropogenic influences like TWW demand targeted, location-specific interventions tailored to local vulnerabilities where natural and human factors intersect.

These findings provide a robust evidence base for targeted salinity management strategies, emphasizing the primacy of protecting areas vulnerable to natural saline sources while implementing careful regulation of drilling practices and treated wastewater reuse. The results underscore that effective salinity control requires integrated approaches addressing both the dominant natural drivers and measurable human contributions to groundwater quality degradation.

## 4. Conclusion

Salinization is a major cause of Israel's water crisis, primarily driven by natural processes that are accelerated by human activity. Over-pumping groundwater disturbs the natural balance, mobilizing deep-seated saline waters and brines. This deteriorates freshwater aquifers, reduces available drinking water, and creates a cycle where new resources are developed only to face the same threat. The problem is worsened by a lack of sufficient knowledge about these saline water bodies, highlighting a critical need for more research and monitoring. This study employs a three-stage framework to predict and explain rising groundwater salinity in Israel. It first identifies causal factors using double machine learning (DML), then predicts chloride concentrations with multiple ML models, and finally explains those predictions using Explainable AI (XAI) methods. DML quantifies the independent influence of each feature on the target variable, while feature selection methods like SHAP and RFE quantify the value that each feature adds to predictive power with respect to the other features. DML isolates the most robust independent effects linked to local hydrological functioning and climatic energy constraints (TWI and temperature), whereas XAI based SHAP and global sensitivity analyses (GSA) emphasize variables that are highly informative for national-scale prediction because they capture hydroclimatic zoning and proximity to salinity sources (precipitation and distance metrics). Apart from this national level study, we are in further process of developing basin level models using the similar framework.

Based on the above analysis our priority interventions for groundwater salinity control can include actions like (a) manage areas near saline bodies, (b) protect riverine areas for freshwater recharge, (c) regulate drilling depths in vulnerable aquifers, (d) enhance precipitation harvesting and recharge and (e) monitor and treat TWW quality for irrigation.

**CRediT authorship contribution statement**

**Laxmi Pandey:** Data preprocessing, Writing-original draft, Writing-review & editing, Coding, Methodology, Conceptualization. **Ariel Meroz:** Data acquisition and curation. **Ben Cheng:** Coding, Methodology. **Ankita Manekar:** Writing-review & editing. **Abhijit Mukherjee:** Writing-review & editing. **Meirav Cohen:** Writing-review & editing,

Supervision, Methodology, Conceptualization and Funding acquisition. **Adway Mitra:** Writing-review & editing, Supervision, Methodology, Conceptualization and Funding acquisition.

**Declaration of competing interest**

The authors declare that they have no known competing financial interests or personal relationships that could have appeared to influence the work reported in this paper.

**Data availability**

Data used in this study were obtained from multiple national and regional sources and are subject to data-use restrictions. The processed datasets supporting the conclusions of this article are available from the corresponding author upon reasonable request, subject to permission from original data providers.

**Acknowledgement**

We acknowledge the Department of Science and Technology, Government of India with Award No. (DST/INT/ISR/P-42/2023) and Israel Ministry of Sciences with Award No. (0006307) for providing financial support to this research under the Indo-Israel Joint Research program. We also acknowledge the use of Generative AI tools for paraphrasing the sentences for enhanced readability and confirm that all the information provided is cross verified by the authors.

**References**

Abanyie, S.K., Apea, O.B., Abagale, S.A., Amuah, E.E.Y. and Sunkari, E.D., 2023. Sources and factors influencing groundwater quality and associated health implications: A review. Emerging Contaminants. 9(2), 100207.

Abid, K., Zouari, K., Dulinski, M., Chkir, N. and Abidi, B., 2011. Hydrologic and geologic factors controlling groundwater geochemistry in the Turonian aquifer (southern Tunisia). Hydrogeology Journal. 19(2), 415-427.

Alagha, J.S., Seyam, M., Md Said, M.A. and Mogheir, Y., 2017. Integrando uma abordagem de inteligência artificial com clusterização por k-means para modelar a salinidade das águas subterrâneas: o caso de um aquífero costeiro de Gaza (Palestina). Hydrogeology Journal. 25, 2347-2361.

Amuah, E., Antwi, K. and Ankamah, N., 2023. Elemental burdens of cosmetics and associated health and environmental impacts: A global view. World J. Adv. Res. Reviews. 20, 76-96.

Avisar, D. and Ronen-Eliraz, G., 2019. Agricultural Irrigation with Effluent–What should we be Worried About?. Journal of Basic & Applied Sciences. 15, 32-39.

Banerjee, P., Singh, V.S., Chatttopadhyay, K., Chandra, P.C. and Singh, B., 2011. Artificial neural network model as a potential alternative for groundwater salinity forecasting. Journal of Hydrology. 398(3-4), 212-220.

Bekkam, V.R., Vajja, V., Nune, R. and Gaur, A., 2013. Estimation and analysis of return flows: Case study. Journal of Hydrologic Engineering. 18(10), 1282-1288.

Bhakar, P. and Singh, A.P., 2019. Groundwater quality assessment in a hyper-arid region of Rajasthan, India. Natural Resources Research. 28(2), 505-522.

Boudibi, S., Fadlaoui, H., Hiouani, F., Bouzidi, N., Aissaoui, A. and Khomri, Z.E., 2024. Groundwater salinity modeling and mapping using machine learning approaches: a case study in Sidi Okba region, Algeria. Environmental Science and Pollution Research. 31(36), 48955-48971.

Bear, J. and Cheng, A.H.D., 2010. Modeling groundwater flow and contaminant transport (Vol. 23, p. 834). Dordrecht: Springer.

Bhanja, S.N., Malakar, P., Mukherjee, A., Rodell, M., Mitra, P. and Sarkar, S., 2019. Using satellite-based vegetation cover as indicator of groundwater storage in natural vegetation areas. Geophysical Research Letters. 46(14), 8082-8092.

Bhattarai, N., Pollack, A., Lobell, D.B., Fishman, R., Singh, B., Dar, A. and Jain, M., 2021. The impact of groundwater depletion on agricultural production in India. Environmental Research Letters. 16(8), 085003.

Bramm, A.M., Matrenin, P.V. and Khalyasmaa, A.I., 2025. A review of XAI methods applications in forecasting runoff and water level hydrological tasks. Mathematics. 13(17), 2830.

Breiman, L., 2001. Random forests. Machine learning. 45(1), 5-32.

Burg, A., Gavrieli, I. and Guttman, J., 2017. Concurrent salinization and development of anoxic conditions in a confined aquifer, Southern Israel. Groundwater. 55(2), 183-198.

Carrard, N., Foster, T. and Willetts, J., 2019. Groundwater as a source of drinking water in southeast Asia and the Pacific: A multi-country review of current reliance and resource concerns. Water. 11(8), 605.

Chakraborty, M., Sarkar, S., Mukherjee, A., Shamsudduha, M., Ahmed, K.M., Bhattacharya, A. and Mitra, A., 2020. Modeling regional-scale groundwater arsenic hazard in the transboundary Ganges River Delta, India and Bangladesh: Infusing physically-based model with machine learning. Science of the total environment. 748, 141107.

Nordin, N.F.C., Mohd, N.S., Koting, S., Ismail, Z., Sherif, M. and El-Shafie, A., 2021. Groundwater quality forecasting modelling using artificial intelligence: A review. Groundwater for Sustainable Development. 14, 100643.

Chefetz, B., Mualem, T. and Ben-Ari, J., 2008. Sorption and mobility of pharmaceutical compounds in soil irrigated with reclaimed wastewater. Chemosphere. 73(8), 1335-1343.

Chen, T., 2016. XGBoost: A Scalable Tree Boosting System. Cornell University.

Chenoweth, J., Hadjinicolaou, P., Bruggeman, A., Lelieveld, J., Levin, Z., Lange, M.A., Xoplaki, E. and Hadjikakou, M., 2011. Impact of climate change on the water resources of the eastern Mediterranean and Middle East region: Modeled 21st century changes and implications. Water Resources Research. 47(6).

Chernozhukov, V., Chetverikov, D., Demirer, M., Duflo, E., Hansen, C., Newey, W. and Robins, J., 2016. Double/debiased machine learning for treatment and causal parameters. arXiv preprint arXiv:1608.00060.

Cohrs, K.H., Varando, G., Carvalhais, N., Reichstein, M. and Camps-Valls, G., 2024. Causal hybrid modeling with double machine learning. *arXiv preprint arXiv:2402.13332*.

Corwin, D.L., Rhoades, J.D. and Šimůnek, J., 2007. Leaching requirement for soil salinity control: Steady-state versus transient models. agricultural water management. 90(3), 165-180.

Delsman, J.R., Hu-A-Ng, K.R.M., Vos, P.C., De Louw, P.G., Oude Essink, G.H., Stuyfzand, P.J. and Bierkens, M.F., 2014. Paleo-modeling of coastal saltwater intrusion during the Holocene: an application to the Netherlands. Hydrology and Earth System Sciences. 18(10), 3891-3905.

Dhurandhar, A., Chen, P.Y., Luss, R., Tu, C.C., Ting, P., Shanmugam, K. and Das, P., 2018. Explanations based on the missing: Towards contrastive explanations with pertinent negatives. Advances in neural information processing systems, 31.

Dorogush, A.V., Ershov, V. and Gulin, A., 2018. CatBoost: gradient boosting with categorical features support. arXiv preprint arXiv:1810.11363.

Duttagupta, S., Bhattacharya, A., Mukherjee, A., Chattopadhyay, S., Bhanja, S.N., Sarkar, S., Malakar, P. and Bhattacharya, J., 2019. Groundwater faecal pollution observation in parts of Indo-Ganges–Brahmaputra river basin from in-situ measurements and satellite-based observations. Journal of Earth System Science. 128(3), 44.

Earman, S. and Dettinger, M., 2011. Potential impacts of climate change on groundwater resources–a global review. Journal of water and climate change. 2(4), 213-229.

Ehteram, M., Kalantari, Z., Ferreira, C.S., Chau, K.W. and Emami, S.M.K., 2022. Prediction of future groundwater levels under representative concentration pathway scenarios using an inclusive multiple model coupled with artificial neural networks. Journal of water and Climate Change. 13(10), 3620-3643.

Faghih, Z., Haroon, A., Jegen, M., Berndt, C., Micallef, A., Mountjoy, J., Schwalenberg, K., Gehrmann, R., Dettmer, J. and Weymer, B.A., 2024, September. Exploring for Offshore Freshened Groundwater: Integrated Geophysical Field Studies. In OCEANS 2024-Halifax (pp. 1-8). IEEE.

Famiglietti, J.S., 2014. The global groundwater crisis. Nature climate change. 4(11), 945-948.

Fetter, C. W., 2001. Applied Hydrogeology (4th ed.). Englewood Cliffs, NJ: Prentice-Hall.

Foster, S. and MacDonald, A., 2021. Groundwater management. Handbook of Catchment Management 2e. 125-152.

Francom, D. and Nachtsheim, A., 2025. A Review and Comparison of Different Sensitivity Analysis Techniques in Practice. arXiv preprint arXiv:2506.11471.

Fuhr, J., Berens, P. and Papies, D., 2024. Estimating Causal Effects with Double Machine Learning--A Method Evaluation. arXiv preprint arXiv:2403.14385.

Ghaffarian, S., Taghikhah, F.R. and Maier, H.R., 2023. Explainable artificial intelligence in disaster risk management: Achievements and prospective futures. International Journal of Disaster Risk Reduction. 98, 104123.

Ghoto, S.M., Abbasi, H., Memon, S.A., Brohi, K.M., Chhachhar, R., Ghanghlo, A.A. and Tunio, I.A., 2025. Multivariate hydrochemical assessment of groundwater quality for irrigation purposes and identifying soil interaction effects and dynamics of NDVI and urban development in alluvial region. Environmental Advances. 19, 100621.

Goyal, M.K. and Surampalli, R.Y., 2018. Impact of climate change on water resources in India. Journal of Environmental Engineering. 144(7), 04018054.

Grant, S.B., Rippy, M.A., Birkland, T.A., Schenk, T., Rowles, K., Misra, S., Aminpour, P., Kaushal, S., Vikesland, P., Berglund, E. and Gomez-Velez, J.D., 2022. Can common pool resource theory catalyze stakeholder-driven solutions to the freshwater salinization syndrome?. Environmental science & technology. 56(19), 13517-13527.

Grünenbaum, N., Greskowiak, J., Sültenfuß, J. and Massmann, G., 2020. Groundwater flow and residence times below a meso-tidal high-energy beach: a model-based analyses of salinity patterns and 3H-3He groundwater ages. Journal of Hydrology. 587, 124948.

Gurmessa, S.K., MacAllister, D.J., White, D., Ouedraogo, I., Lapworth, D. and MacDonald, A., 2022. Assessing groundwater salinity across Africa. Science of the Total Environment. 828, 154283.

Hassani, A., Azapagic, A. and Shokri, N., 2021. Global predictions of primary soil salinization under changing climate in the 21st century. Nature communications. 12(1), 6663.

Hashem, M.S. and Qi, X., 2021. Treated wastewater irrigation—A review. Water. 13(11), 1527.

Hong, Y., Zhu, Z., Liao, W., Yan, Z., Feng, C. and Xu, D., 2023. Freshwater water-quality criteria for chloride and guidance for the revision of the water-quality standard in China. International journal of environmental research and public health. 20(4), 2875.

Howard, G., Calow, R., Macdonald, A. and Bartram, J., 2016. Climate change and water and sanitation: likely impacts and emerging trends for action. Annual review of environment and resources. 41(1), 253-276.

Iooss, B. and Lemaître, P., 2015. A review on global sensitivity analysis methods. Uncertainty management in simulation-optimization of complex systems: algorithms and applications. 101-122.

*Israel, C. of., 2021. Prevention of water resources contamination and salinization, monitoring and restoration.

Jacovi, A., Swayamdipta, S., Ravfogel, S., Elazar, Y., Choi, Y. and Goldberg, Y., 2021. Contrastive explanations for model interpretability. arXiv preprint arXiv:2103.01378.

Jiang, S., Zheng, Y., Wang, C. and Babovic, V., 2022. Uncovering flooding mechanisms across the contiguous United States through interpretive deep learning on representative catchments. Water Resources Research. 58(1), e2021WR030185.

Jung, Y., Tian, J. and Bareinboim, E., 2021, May. Estimating identifiable causal effects through double machine learning. In Proceedings of the AAAI Conference on Artificial Intelligence (Vol. 35, No. 13, 12113-12122).

Jung, H., Saynisch-Wagner, J. and Schulz, S., 2024. Can eXplainable AI offer a new perspective for groundwater recharge estimation?—Global-scale modeling using neural network. Water Resources Research. 60(4), e2023WR036360.

Khan, S., Xevi, E. and Meyer, W.S., 2003. Salt, water, and groundwater management models to determine sustainable cropping patterns in shallow saline groundwater regions of Australia. Journal of crop production. 7(1-2), 325-340.

Kellenberger, B., Tasar, O., Bhushan Damodaran, B., Courty, N. and Tuia, D., 2021. Deep domain adaptation in earth observation. Deep Learning for the Earth Sciences: A Comprehensive Approach to Remote Sensing, Climate Science, and Geosciences. 90-104.

King, J., Mulder, T., Oude Essink, G. and Bierkens, M.F., 2022. Joint estimation of groundwater salinity and hydrogeological parameters using variable-density groundwater flow, salt transport modelling and airborne electromagnetic surveys. Advances in Water Resources. 160, 104118.

Klapp, I., Korach-Rechtman, H., Kurtzman, D., Levy, G., Maffettone, R., Malato, S., Manaia, C., Manoli, K., Moshe, O., Rimelman, A. and Rizzo, L., 2023. Mitigating risks and maximizing sustainability of treated wastewater reuse for irrigation.

Kratzert, F., Klotz, D., Brenner, C., Schulz, K. and Herrnegger, M., 2018. Rainfall–runoff modelling using long short-term memory (LSTM) networks. Hydrology and Earth System Sciences, 22(11), 6005-6022.

Lal, A. and Datta, B., 2018, November. Genetic programming and gaussian process regression models for groundwater salinity prediction: Machine learning for sustainable water resources management. In 2018 IEEE conference on technologies for sustainability (SusTech) (1-7). IEEE.

Larsen, F., Tran, L.V., Van Hoang, H., Tran, L.T., Christiansen, A.V. and Pham, N.Q., 2017. Groundwater salinity influenced by Holocene seawater trapped in incised valleys in the Red River delta plain. Nature Geoscience. 10(5), 376-381.

Lundberg, S.M. and Lee, S.I., 2017. A unified approach to interpreting model predictions. Advances in neural information processing systems. 30.

Li, Z.L., Leng, P., Zhou, C., Chen, K.S., Zhou, F.C. and Shang, G.F., 2021. Soil moisture retrieval from remote sensing measurements: Current knowledge and directions for the future. Earth-Science Reviews, 218, p.103673.

Linardatos, P., Papastefanopoulos, V., & Kotsiantis, S. (2020). Explainable ai: A review of machine learning interpretability methods. Entropy. 23(1), 18.

Maier, H.R., Taghikhah, F.R., Nabavi, E., Razavi, S., Gupta, H., Wu, W., Radford, D.A. and Huang, J., 2024. How much X is in XAI: Responsible use of "Explainable" artificial intelligence in hydrology and water resources. Journal of Hydrology X. 25, 100185.

Malakar, P., Mukherjee, A., Bhanja, S.N., Ganguly, A.R., Ray, R.K., Zahid, A., Sarkar, S., Saha, D. and Chattopadhyay, S., 2021. Three decades of depth-dependent groundwater response to climate variability and human regime in the transboundary Indus-Ganges-Brahmaputra-Meghna mega river basin aquifers. Advances in Water Resources. 149, 103856.

Masciopinto, C., Liso, I.S., Caputo, M.C. and De Carlo, L., 2017. An integrated approach based on numerical modelling and geophysical survey to map groundwater salinity in fractured coastal aquifers. Water. 9(11), 875.

McDonald, R.C., 2016. A process for developing and revising a learning progression on sea level rise using learners' explanations (Doctoral dissertation, University of Maryland, College Park).

McCleskey, R.B., Cravotta III, C.A., Miller, M.P., Tillman, F., Stackelberg, P., Knierim, K.J. and Wise, D.R., 2023. Salinity and total dissolved solids measurements for natural waters: An overview and a new salinity method based on specific conductance and water type. Applied Geochemistry. 154, 105684.

McNeill, J.D., 1988, March. Advances in electromagnetic methods for groundwater studies. In 1st EEGS Symposium on the Application of Geophysics to Engineering and Environmental Problems (cp-214). European Association of Geoscientists & Engineers.

Meles, M.B., Younger, S.E., Jackson, C.R., Du, E. and Drover, D., 2020. Wetness index based on landscape position and topography (WILT): Modifying TWI to reflect landscape position. Journal of environmental management. 255, 109863.

Mohanavelu, A., Naganna, S.R. and Al-Ansari, N., 2021. Irrigation induced salinity and sodicity hazards on soil and groundwater: An overview of its causes, impacts and mitigation strategies. Agriculture. 11(10), 983.

Molnar, C., 2020. Interpretable machine learning. Lulu. com.

Morris, M.D., 1991. Factorial sampling plans for preliminary computational experiments. Technometrics. 33(2), 161-174.

Mosaffa, M., Nazif, S., Amirhosseini, Y.K., Balderer, W. and Meiman, H.M., 2021. An investigation of the source of salinity in groundwater using stable isotope tracers and GIS: a case study of the Urmia Lake basin, Iran. Groundwater for Sustainable Development. 12, 100513.

Mosavi, A., Hosseini, F.S., Choubin, B., Goodarzi, M. and Dineva, A.A., 2020. Groundwater salinity susceptibility mapping using classifier ensemble and Bayesian machine learning models. Ieee Access. 8, 145564-145576.

Mosavi, A., Sajedi Hosseini, F., Choubin, B., Taromideh, F., Ghodsi, M., Nazari, B. and Dineva, A.A., 2021. Susceptibility mapping of groundwater salinity using machine learning models. Environmental Science and Pollution Research. 28(9), 10804-10817.

Mothilal, R.K., Sharma, A. and Tan, C., 2020, January. Explaining machine learning classifiers through diverse counterfactual explanations. In Proceedings of the 2020 conference on fairness, accountability, and transparency (607-617).

Mondal, N.C., Adike, S., Anand Raj, P., Singh, V.S., Ahmed, S. and Jayakumar, K.V., 2018. Assessing aquifer vulnerability using GIS-based DRASTIC model coupling with hydrochemical parameters in hard rock area from Southern India. In Groundwater: select proceedings of ICWEES-2016 (67-82). Singapore: Springer Singapore.

Mukherjee, A., Scanlon, B.R., Aureli, A., Langan, S., Guo, H. and McKenzie, A.A. eds., 2020. Global groundwater: source, scarcity, sustainability, security, and solutions. Elsevier.

Mukherjee, A., Sarkar, S., Chakraborty, M., Duttagupta, S., Bhattacharya, A., Saha, D., Bhattacharya, P., Mitra, A. and Gupta, S., 2021. Occurrence, predictors and hazards of elevated groundwater arsenic across India through field observations and regional-scale AI-based modeling. Science of the Total Environment. 759, 143511.

Mullainathan, S. and Spiess, J., 2017. Machine learning: an applied econometric approach. Journal of Economic Perspectives, 31(2), 87-106.

Mushtaq, H., Akhtar, T., Masood, A. and Saeed, F., 2024. Hydrologic interpretation of machine learning models for 10-daily streamflow simulation in climate sensitive upper Indus catchments. Theoretical & Applied Climatology. 155(6).

Naser, A.M., Unicomb, L., Doza, S., Ahmed, K.M., Rahman, M., Uddin, M.N., Quraishi, S.B., Selim, S., Shamsudduha, M., Burgess, W. and Chang, H.H., 2017. Stepped-wedge cluster-randomised controlled trial to assess the cardiovascular health effects of a managed aquifer recharge initiative to reduce drinking water salinity in southwest coastal Bangladesh: study design and rationale. BMJ open. 7(9), e015205.

Nathan, N.S., Saravanane, R. and Sundararajan, T., 2017. Application of ANN and MLR models on groundwater quality using CWQI at Lawspet, Puducherry in India. Journal of Geoscience and Environment Protection. 5(03), 99.

Nosair, A.M., Shams, M.Y., AbouElmagd, L.M., Hassanein, A.E., Fryar, A.E. and Abu Salem, H.S., 2022. Predictive model for progressive salinization in a coastal aquifer using artificial intelligence and hydrogeochemical techniques: A case study of the Nile Delta aquifer, Egypt. Environmental Science and Pollution Research. 29(6), 9318-9340.

Oren, O., Yechieli, Y., Böhlke, J.K. and Dody, A., 2004. Contamination of groundwater under cultivated fields in an arid environment, central Arava Valley, Israel. Journal of Hydrology. 290(3-4), 312-328.

Panigrahi, B., Razavi, S., Doig, L.E., Cordell, B., Gupta, H.V. and Liber, K., 2025. On robustness of the explanatory power of machine learning models: Insights from a new explainable AI approach using sensitivity analysis. Water Resources Research. 61(3), e2024WR037398.

Pedregosa, F., Varoquaux, G., Gramfort, A., Michel, V., Thirion, B., Grisel, O., Blondel, M., Prettenhofer, P., Weiss, R., Dubourg, V. and Vanderplas, J., 2011. Scikit-learn: Machine learning in Python. the Journal of machine Learning research. 12, 2825-2830.

Pradhan, B., Lee, S., Dikshit, A. and Kim, H., 2023. Spatial flood susceptibility mapping using an explainable artificial intelligence (XAI) model. Geoscience Frontiers. 14(6), 101625.
46

Pratiwi, O.A., Nurwidyanto, M.I. and Widada, S., 2024. Analysis of Groundwater Salinity Distribution Based on Electrical Conductivity (EC) and Hydrochemical Approaches to Deep Wells in Sayung Subdistrict, Demak Central Java. Cognizance Journal of Multidisciplinary Studies. 4(10), 53-61.

Raheja, H., Goel, A. and Pal, M., 2022. Prediction of groundwater quality indices using machine learning algorithms. Water Practice & Technology. 17(1), 336-351.

Reichstein, M., Camps-Valls, G., Stevens, B., Jung, M., Denzler, J., Carvalhais, N. and Prabhat, F., 2019. Deep learning and process understanding for data-driven Earth system science. Nature. 566(7743), 195-204.

Ribeiro, M.T., Singh, S. and Guestrin, C., 2016. Model-agnostic interpretability of machine learning. arXiv preprint arXiv:1606.05386.

Ribeiro, M.T., Singh, S. and Guestrin, C., 2018, April. Anchors: High-precision model-agnostic explanations. In Proceedings of the AAAI conference on artificial intelligence (Vol. 32, No. 1).

Rosenthal, E., Vinokurov, A., Ronen, D., Magaritz, M. and Moshkovitz, S., 1992. Anthropogenically induced salinization of groundwater: A case study from the Coastal Plain aquifer of Israel. Journal of Contaminant Hydrology. 11(1-2), 149-171.

Roshani, A. and Hamidi, M., 2022. Groundwater level fluctuations in coastal aquifer: Using artificial neural networks to predict the impacts of climatical CMIP6 scenarios. Water Resources Management. 36(11), 3981-4001.

Rosinger, A.Y., McGrosky, A., Jacobson, H., Hinz, E., Sadhir, S., Wambua, F., Otube, T., Baker, L.J., Sherwood, A.C., Chrissy-Mbeng, T. and Broyles, L.M., 2025. Drinking water NaCl is associated with hypertension and albuminuria: a panel study. Hypertension. 82(8), 368-1378.

Roy, D.K., Sarkar, T.K., Munmun, T.H., Paul, C.R. and Datta, B., 2025. A review on the applications of machine learning and deep learning to groundwater salinity modeling: Present status, challenges, and future directions. Discover Water. 5(1), 16.

Sahour, H., Gholami, V. and Vazifedan, M., 2020. A comparative analysis of statistical and machine learning techniques for mapping the spatial distribution of groundwater salinity in a coastal aquifer. Journal of Hydrology. 591, 125321.

Sahour, S., Khanbeyki, M., Gholami, V., Sahour, H., Kahvazade, I. and Karimi, H., 2023. Evaluation of machine learning algorithms for groundwater quality modeling. Environmental Science and Pollution Research. 30(16), 46004-46021.

Saltelli, A., Tarantola, S., Campolongo, F. and Ratto, M., 2004. Sensitivity analysis in practice: a guide to assessing scientific models (Vol. 1). New York: Wiley.

Samek, W., Montavon, G., Lapuschkin, S., Anders, C.J. and Müller, K.R., 2021. Explaining deep neural networks and beyond: A review of methods and applications. Proceedings of the IEEE. 109(3), 247-278.

Sarkar, S., Das, K. and Mukherjee, A., 2024. Groundwater salinity across India: predicting occurrences and controls by field-observations and machine learning modeling. Environmental Science & Technology. 58(8), 3953-3965.

Schafer, D., Sun, J., Jamieson, J., Siade, A.J., Atteia, O. and Prommer, H., 2020. Model-based analysis of reactive transport processes governing fluoride and phosphate release and attenuation during managed aquifer recharge. Environmental Science & Technology. 54(5), 2800-2811.

Schiavo, M., Giambastiani, B.M., Greggio, N., Colombani, N. and Mastrocicco, M., 2024. Geostatistical assessment of groundwater arsenic contamination in the Padana Plain. Science of the Total Environment. 931, 172998.

Schiller, J., Stiller, S. and Ryo, M., 2025. Artificial intelligence in environmental and Earth system sciences: explainability and trustworthiness. Artificial Intelligence Review. 58(10), 316.

Shapiro, A.M. and Day-Lewis, F.D., 2022. Reframing groundwater hydrology as a data-driven science. Groundwater. 60(4), 455-456.

Shen, C., Appling, A.P., Gentine, P., Bandai, T., Gupta, H., Tartakovsky, A., Baity-Jesi, M., Fenicia, F., Kifer, D., Li, L. and Liu, X., 2023. Differentiable modelling to unify machine learning and physical models for geosciences. Nature Reviews Earth & Environment. 4(8), 552-567.

Shiri, N., Shiri, J., Yaseen, Z.M., Kim, S., Chung, I.M., Nourani, V. and Zounemat-Kermani, M., 2021. Development of artificial intelligence models for well groundwater quality simulation: Different modeling scenarios. Plos one. 16(5), e0251510.

Shtull-Trauring, E., Cohen, A., Ben-Hur, M., Israeli, M. and Bernstein, N., 2022. NPK in treated wastewater irrigation: Regional scale indices to minimize environmental pollution and optimize crop nutritional supply. Science of The Total Environment. 806, 150387.

Shmueli, G., 2010. To explain or to predict?. Statistical science. 289-310.

Sobol, I.M., 2001. Global sensitivity indices for nonlinear mathematical models and their Monte Carlo estimates. Mathematics and computers in simulation. 55(1-3), 271-280.

Van Stein, B., Raponi, E., Sadeghi, Z., Bouman, N., Van Ham, R.C. and Bäck, T., 2022. A comparison of global sensitivity analysis methods for explainable AI with an application in genomic prediction. IEEE Access. 10, 103364-103381.

Sun, A.Y., Scanlon, B.R., Zhang, Z., Walling, D., Bhanja, S.N., Mukherjee, A. and Zhong, Z., 2019. Combining physically based modeling and deep learning for fusing GRACE satellite data: can we learn from mismatch?. Water Resources Research. 55(2), 1179-1195.

Sundararajan, M., Taly, A. and Yan, Q., 2017, July. Axiomatic attribution for deep networks. In International conference on machine learning (3319-3328). PMLR.

Tahmasebi, P., Kamrava, S., Bai, T. and Sahimi, M., 2020. Machine learning in geo- and environmental sciences: From small to large scale. Advances in Water Resources. 142, 103619.

Thorslund, J. and van Vliet, M.T., 2020. A global dataset of surface water and groundwater salinity measurements from 1980–2019. Scientific Data. 7(1), 231.

Topp, S.N., Barclay, J., Diaz, J., Sun, A.Y., Jia, X., Lu, D., Sadler, J.M. and Appling, A.P., 2023. Stream temperature prediction in a shifting environment: Explaining the influence of deep learning architecture. Water Resources Research. 59(4), e2022WR033880.

Tiwari, V. M., Wahr, J., & Swenson, S., 2009. Dwindling groundwater resources in northern India, from satellite gravity observations. Geophysical Research Letters, 36(18).

Morelos-Villegas, A., Condal, A.R. and Ardisson, P.L., 2018. Spatial heterogeneity and seasonal structure of physical factors and benthic species in a tropical coastal lagoon, Celestun, Yucatan Peninsula. Regional Studies in Marine Science. 22, 136-146.

Vengosh, A. and Rosenthal, E., 1994. Saline groundwater in Israel: its bearing on the water crisis in the country. Journal of Hydrology. 156(1-4), 389-430.

Vineis, P., Chan, Q. and Khan, A., 2011. Climate change impacts on water salinity and health. Journal of epidemiology and global health. 1(1), 5-10.

Wang, S., Peng, H., Hu, Q. and Jiang, M., 2022. Analysis of runoff generation driving factors based on hydrological model and interpretable machine learning method. Journal of Hydrology: Regional Studies. 42, 101139.

Wang, Y., Wang, W., Ma, Z., Zhao, M., Li, W., Hou, X., Li, J., Ye, F. and Ma, W., 2023. A deep learning approach based on physical constraints for predicting soil moisture in unsaturated zones. Water Resources Research. 59(11), e2023WR035194.

Wachter, S., Mittelstadt, B. and Russell, C., 2017. Counterfactual explanations without opening the black box: Automated decisions and the GDPR. Harv. JL & Tech., 31, 841.

Wu, J., Wang, Z., Dong, J., Cui, X., Tao, S. and Chen, X., 2023. Robust runoff prediction with explainable artificial intelligence and meteorological variables from deep learning ensemble model. Water Resources Research. 59(9), e2023WR035676.

Wunsch, A., Liesch, T. and Broda, S., 2022. Deep learning shows declining groundwater levels in Germany until 2100 due to climate change. Nature communications. 13(1), 1221.

Ye, S., Li, J., Chai, Y., Liu, L., Sivapalan, M. and Ran, Q., 2025. Using explainable artificial intelligence (XAI) as a diagnostic tool: An application for deducing hydrologic connectivity at watershed scale. arXiv e-prints, pp.arXiv-2509.

Yechieli, Y., Starinsky, A. and Rosenthal, E., 1992. Evolution of brackish groundater in a typical arid region: Northern Arava Rift Valley, southern Israel. Applied Geochemistry. 7(4), 361-374.

Yechieli, Y. and Sivan, O., 2011. The distribution of saline groundwater and its relation to the hydraulic conditions of aquifers and aquitards: examples from Israel. Hydrogeology Journal. 19(1), 71-81.

Yesilnacar, M.I. and Sahinkaya, E., 2012. Artificial neural network prediction of sulfate and SAR in an unconfined aquifer in southeastern Turkey. Environmental Earth Sciences. 67(4), 1111-1119.

Zhang, Y. and Shen, Y., 2019. Wastewater irrigation: past, present, and future. Wiley Interdisciplinary Reviews: Water. 6(3), e1234.

**APPENDIX A**

Various machine learning based regressions models defined below including LR, RF, XGBoost, ANN, LSTM and CNN, were used for training purpose.

**Linear Regression (LR)**: Linear regression model used mainly for regression purpose, models the relation between a dependent variable and one or more independent variables using a linear function. Assuming a linear relation, model minimizes the error between the actual and predicted values.

**Random Forest (RF):** Random Forest is an ensemble learning method used both for classification and regression tasks. This model used 100 decision trees (estimators) to make predictions thereby reducing overfitting and improving the model performance.

**Extreme Gradient Boosting (XGBoost):** It is a machine learning technique that builds up a strong predictive model by combining many weak models, typically decision trees, in a sequential manner. Each new tree is focused on correcting the errors made by the previous trees thereby improving overall accuracy. It uses a gradient descent technique to minimize the objective function, allowing the model to learn from the errors. This method can be used for both classification and regression tasks.

**Feed Forward Neural Network:** A feedforward neural network is an artificial neural network (ANN) based architecture which is commonly used for tasks like classification and regression. It is designed to model complex relationships in data, where information moves from input to the output layer through one or more hidden layers. The network architecture included fully connected layers with variations of 5, 10, and 15 hidden layers, each of them having 10 neurons activated by ReLU or LeakyReLU functions, culminating in a single linear-activated output neuron for regression. A comprehensive hyperparameter search evaluated four optimizers—Adam, SGD, RMSprop, and Adagrad—across different activation functions and layer depths. Models were trained for 50 epochs and assessed using $R^2$ score and Mean Absolute

Error, with final model selection based on the highest cross-validated R² performance for each optimizer configuration.

**Long Short-Term Memory (LSTM) Model**

Long Short-Term Memory (LSTM) models are recurrent neural network-based models designed mainly for time series prediction. They effectively capture long-term dependencies in sequential data, thereby forecasting future values based on historical trends. During modelling we organized the data into five-timestep sequences grouped by drill locations. Input features and target salinity values were normalized to enhance model convergence, with performance evaluation conducted through 5-fold cross-validation using R² score. The network architecture employed stacked LSTM layers ranging from three to four layers with decreasing units (50, 100, 150), incorporating ReLU and Tanh activation functions alongside regularization through dropout layers (rate=0.3) and batch normalization. The model culminated in fully connected dense layers with a final linear output for regression, with hyperparameter optimization focusing on Adam and RMSprop optimizers across various activation functions and hidden unit configurations.

**Convolutional Neural Networks (CNN) Model**

CNNs are neural models primarily designed for processing grid-like data, such as images or time series. While traditionally used in computer vision, 1D CNNs are highly effective for time-series forecasting due to their ability to extract local patterns and hierarchical features through convolutional filters. The modelling framework organized drill location data into five-timestep sequences to capture temporal dependencies, with all input features standardized and model performance assessed through 5-fold cross-validation based on R² score and Root Mean Square Error. The network architecture featured two Conv1D layers with increasing filters (32→64 or 64→128) and kernel size 3, activated by ReLU or LeakyReLU functions. Comprehensive regularization was implemented through batch normalization after each convolutional and dense layer, combined with strategic dropout (rate=0.2 after convolutional layers, 0.3 after dense layers). The convolutional outputs were flattened and processed through two fully connected layers (100 and 50 units) before final linear regression output, with hyperparameter optimization examining Adam, RMSprop, and Adagrad optimizers across different activation functions and initial filter sizes.

## APPENDIX B

Various Methods mentioned below were used for feature importance analysis.

**RFE-RF** or Recursive feature selection infused with Random Forest method methodically eliminates the features which are least important features based on the performance of the model. This hybrid method trains the RF model on all input features to compute their importance scores depending on their contribution in predicting salinity. It then recursively eliminates the least relevant features in each iteration, retraining the model with the reduced feature set each time. This process continues until a subset of features is found that maximizes the predictive skill (Mosavi et al., 2020; Boudibi et al., 2024).

**GSA** - Sobol based Global Sensitivity Analysis (GSA) is a robust method that estimates how much variance in a model's predictions is attributed to uncertainty in each of the input parameters (Sobol, Morris, 2001). While other local sensitivity methods compute perturbations around a fixed point, this global approach uses Sobol indices to decompose the total output variance into different components attributable to individual parameters and their interactions. Specifically, first-order Sobol indices ($S_1$) measure the individual contribution of each input parameter to the output variance. Meanwhile, total-order indices ($S_T$) capture a parameter's total effect, including all its higher-order interactions with other variables. The method is model-independent, requiring no assumptions of linearity or monotonicity, and relies on Monte Carlo integration typically using efficient, quasi-random Sobol sequences to explore the multi-dimensional input space (Iooss and Lemaître, 2015).

**SHAP** - Shapley Values are a concept in cooperative Game Theory used to compute contributions of each member in a coalition. SHAP is an algorithm for efficient computation of Shapley Values, which requires an accurate predictive model. SHAP quantifies the marginal contribution of each feature to every prediction, ensuring fairness in distributed importance (Lundberg and Lee., 2017; Wang et al., 2022; Mushtaq et al., 2024). It also captures feature interactions and offers model-agnostic interpretability (e.g., via summary plots or dependence plots).

## APPENDIX C

The SHAP summary and Global sensitivity analysis (GSA) plot derived from different machine learning based regression models is showcased below:

**Linear Regression model**

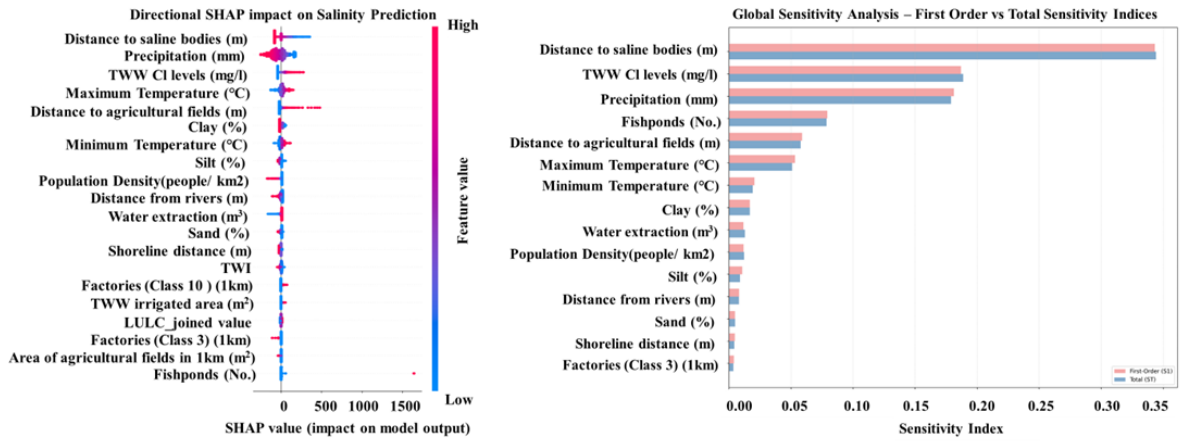

Fig. C.1. SHAP summary (left panel) and Global sensitivity analysis (GSA) plot (right panel) derived from linear regression model.

**Random Forest model**

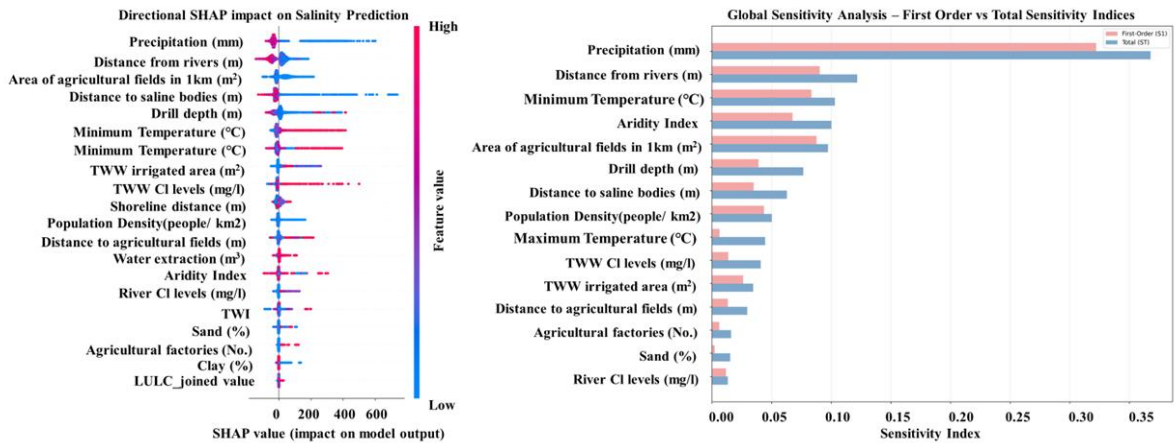

Fig. C.2. SHAP summary (left panel) and Global sensitivity analysis (GSA) plot (right panel) derived from Random Forest model.

**eXtreme Gradient Boost model**

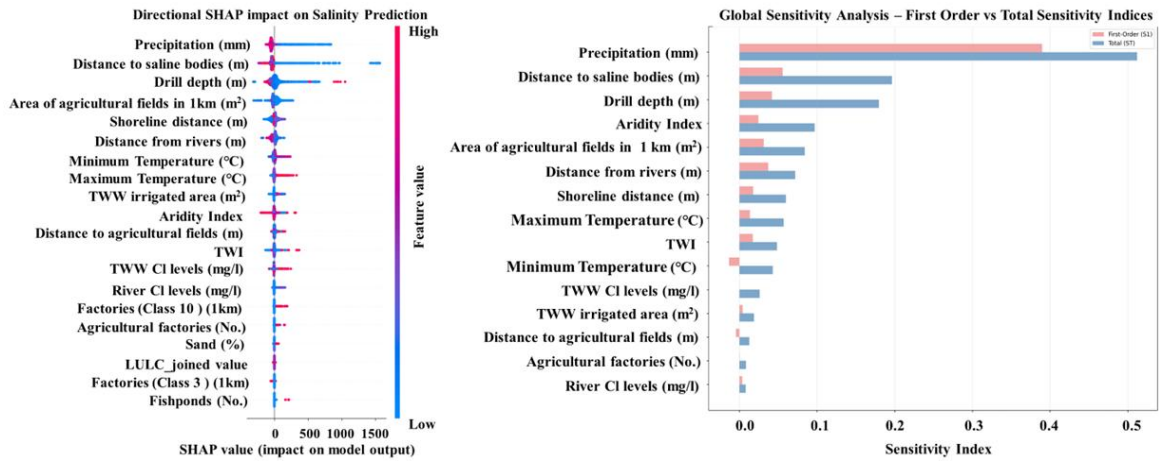

Fig. C.3. SHAP summary (left panel) and Global sensitivity analysis (GSA) plot (right panel) derived from XGBoost model.

**ANN model**

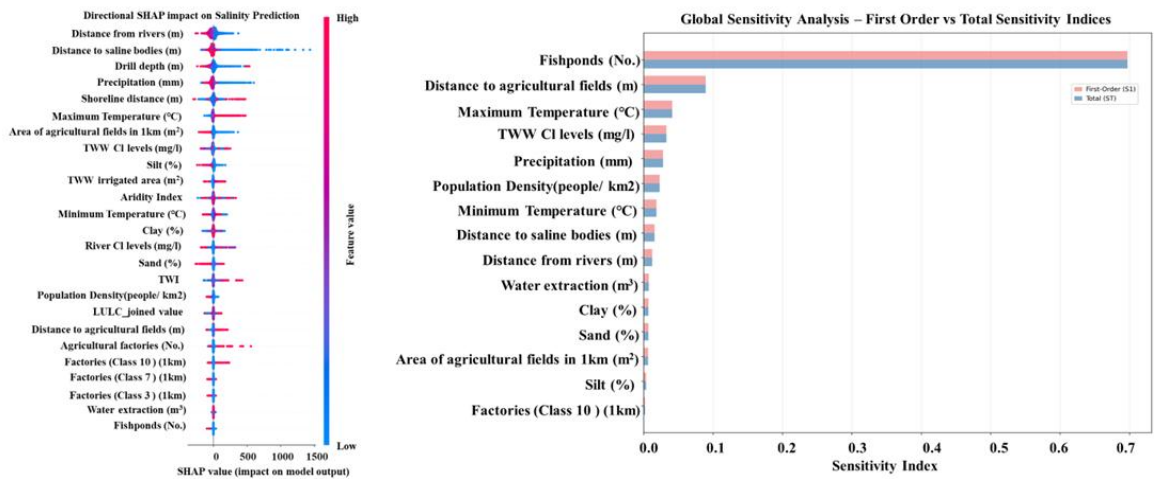

Fig. C.4. SHAP summary (left panel) and Global sensitivity analysis (GSA) plot (right panel) derived from Artificial Neural Network (ANN) model.

**LSTM model**

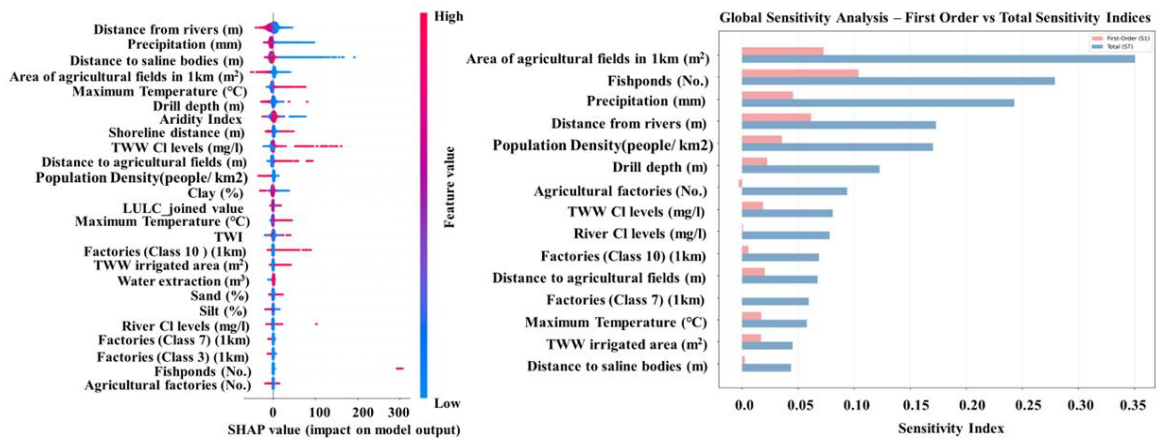

Fig. C.5. SHAP summary (left panel) and Global sensitivity analysis (GSA) plot (right panel) derived from Long-short term Memory (LsTm) model.

**CNN model**

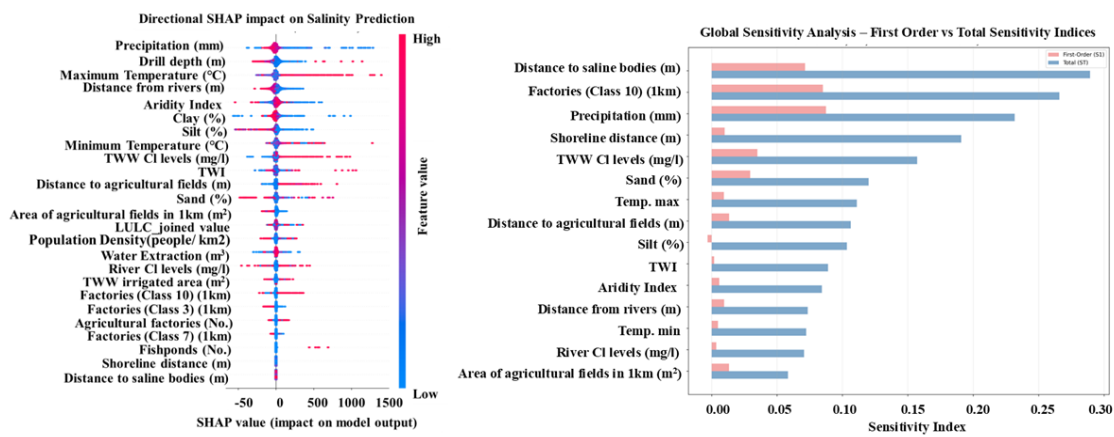

Fig. C.6. SHAP summary (left panel) and Global sensitivity analysis (GSA) plot (right panel) derived from Convolution Neural Network (CNN) model.